%% file: main.tex
\pdfoutput=1
\documentclass{article} 
\usepackage{iclr2027_conference,times}

\usepackage[utf8]{inputenc} 
\usepackage[T1]{fontenc}    
\usepackage{hyperref}       
\usepackage{url}            
\usepackage{booktabs}       
\usepackage{amsfonts}       
\usepackage{nicefrac}       
\usepackage{microtype}      
\usepackage{xcolor}         
\usepackage{graphicx}
\usepackage{multirow} 
\usepackage{amsmath}
\usepackage{caption}
\usepackage{subcaption}
\usepackage{algorithm}
\usepackage{algpseudocode}
\usepackage{stmaryrd} 
\usepackage{etoc}

\input{math_commands.tex}

\title{Learning Directional Randers Geometry for Latent-Space Exploration}

\author{%
  Théau Blanchard \\
  UMR 1346 HeKA \\
  UPCite-INRIA-INSERM \\
  \& GE Healthcare \\ 
  Paris, France \\
  \texttt{theau.blanchard@inria.fr}
  \And
  Stéphanie Allassonnière \\
  UMR 1346 HeKA \\
  UPCite-INRIA-INSERM \\
  Paris, France
}

\def\arxivheadertext{Learning Directional Randers Geometry for Latent-Space Exploration.}
\arxivfinalcopy
\begin{document}

\maketitle

\begin{abstract}
Deep neural networks extract powerful latent representations from images, enabling fast and efficient downstream analysis. However, these latent spaces lack explicit geometric structure, leading to possibly poor interpolation fidelity and misleading extrapolation. Latent interpolation techniques assume symmetric transition costs which are often not the case for real world data. Their symmetric nature cannot capture directional processes, such as disease progression, where certain transitions are impossible (\textit{e.g.} tumor growth dynamics or cognitive improvement in Alzheimer's disease). Although Riemannian metrics extend Euclidean geometry by incorporating curvature, they remain symmetric. We show that understanding asymmetry and not curvature alone directly addresses these problems. To do so we propose a new Finsler-Randers metric learning framework. It incorporates direction-dependent distances, thus enabling the computation of trajectories respecting monotonic sequential constraints. We demonstrate its potential for interpolation and extrapolation on controlled cross-sectional datasets.
\end{abstract}


\section{Introduction}

The progressive switch in image processing from hand-crafted features to features extracted from deep neural networks has significantly improved performance in various fields such as radiology \citep{astaraki2021comparative}, computational pathology \citep{mcgenity2024artificial}, image classification \citep{xian2017zero}, and others. Once equipped with a decoder, latent space exploration serves as a proxy for exploration of the image space. In clinical settings, this approach is motivated by the analysis of patient over time, whether through interpolation or extrapolation of trajectories, or synthesis of longitudinal data. However, these gains come at the cost of interpretability. The latent spaces in which these features live are not naturally equipped with an explicit geometric structure: the assumption of Euclidean geometry falls short when exploring the latent space \citep{mi2021revisiting}. While the lipschitzness of standards networks \citep{virmaux2018lipschitz} enables local understanding of latent neighborhoods, this local view is not enough to characterize the global structure needed for reliable long-range interpolations and extrapolation. 

\begin{figure}[ht]
     \centering
     \begin{subfigure}[b]{0.48\textwidth}
         \vspace{0pt}
         \centering
         \includegraphics[width=\linewidth]{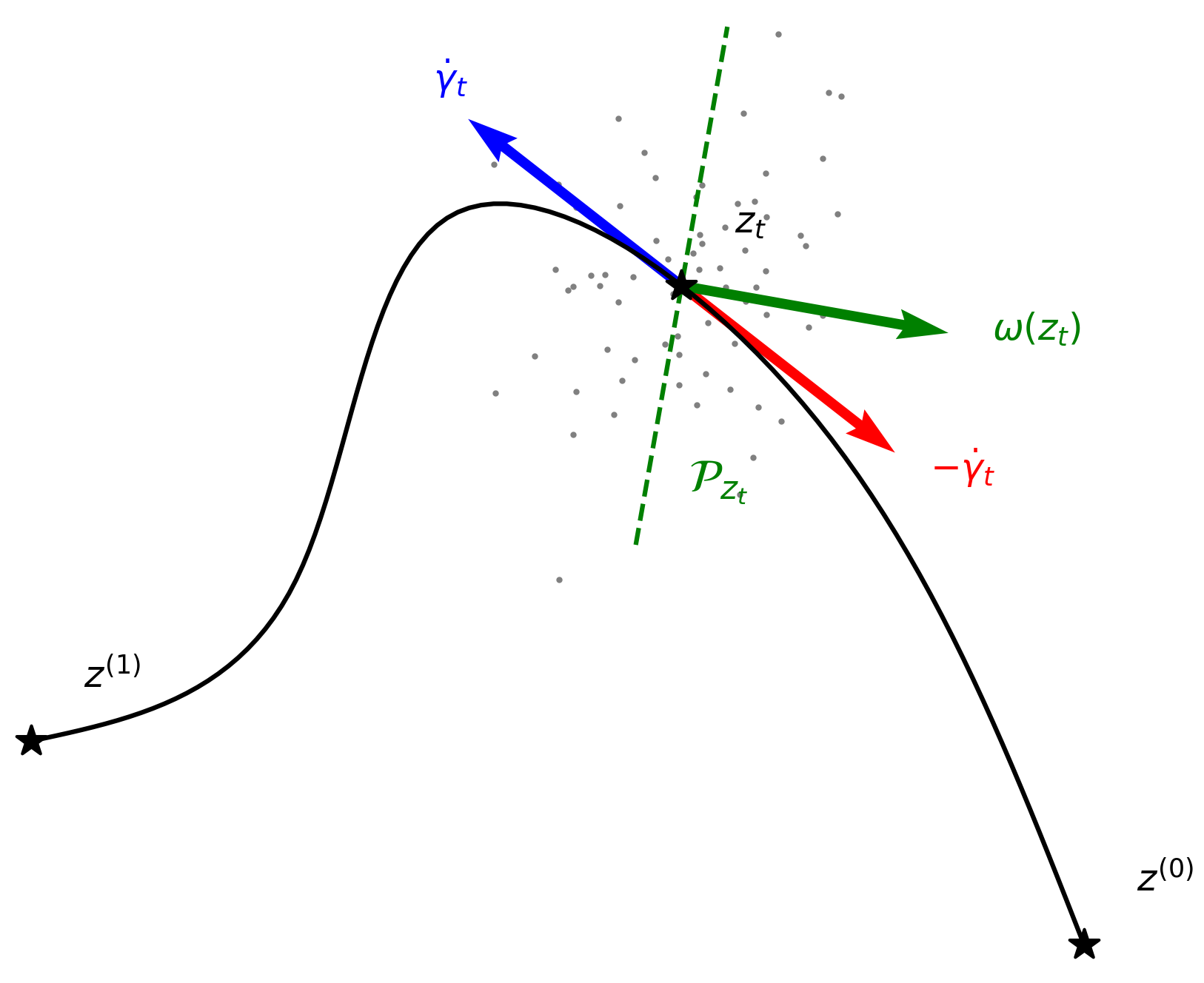}
         \caption{}
         \label{fig:explication_hyperplan}
     \end{subfigure}
     \hfill
     \begin{subfigure}[b]{0.48\textwidth}
        \vspace{0pt}
        \centering     
        \includegraphics[width=\linewidth]{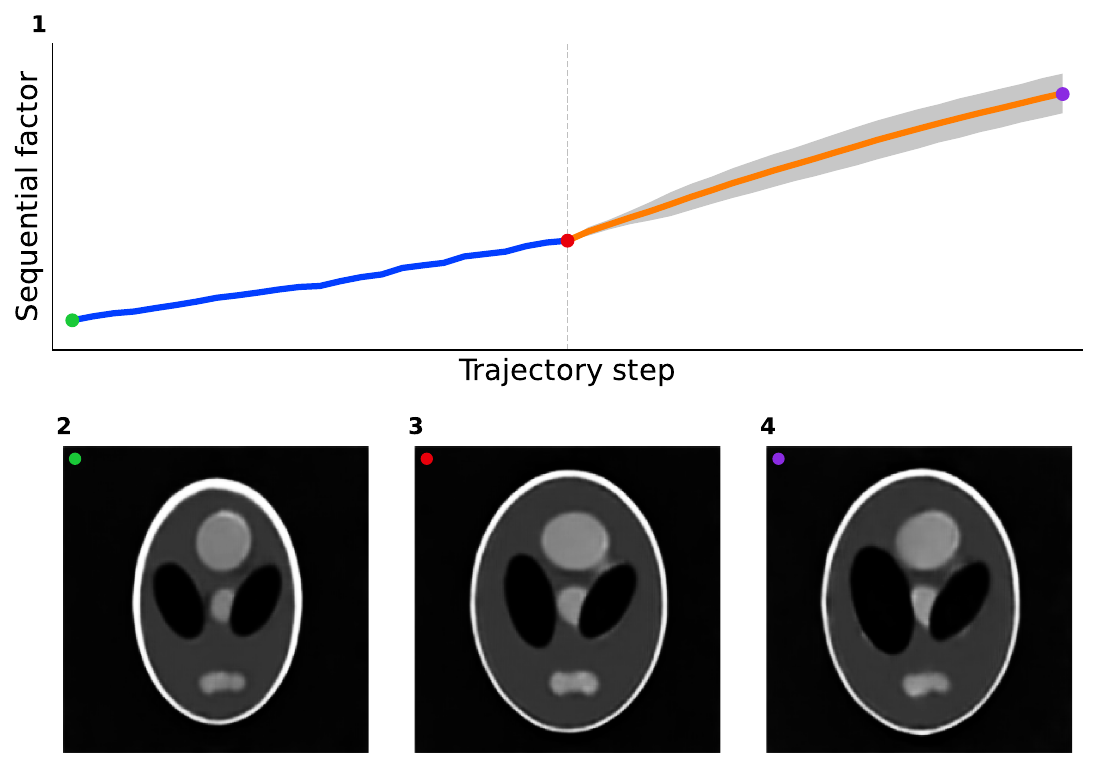}
        \caption{}
        \label{fig:shepplogan_inter_then_extra}
     \end{subfigure}
        \caption{
        \textbf{a.} Example of trajectory $\gamma$ between $z^{0}$ and $z^{1}$. At $\gamma_t =z_t$ we consider the forward and backward trajectory. Because $ \lVert \dot{\gamma}_t\rVert_{G(\gamma_t)} =  \lVert -\dot{\gamma}_t\rVert_{G(\gamma_t)}$, the only difference in energy is induced by $\langle\omega,\,.\,\rangle$.  The angle between $\omega(z_t)$ and $-\dot{\gamma}_t$ is small, so the dot product is positive, hence the associated energy is high. On the contrary, because  $\omega(z_t)$ and $\dot{\gamma}_t$ are almost of opposite direction, the dot product is negative, hence the associated energy is lower : this direction is preferred.
        \textbf{b.} Interpolation then extrapolations. \textit{1} : Sequential factors along the trajectory. In \textit{blue} the interpolated trajectory, in \textit{orange} the mean extrapolated sequential factors with $\pm2\sigma$ on $50$ trajectories. \textit{2} : Start sample, \textit{3} : End of interpolation, \textit{4} : One extrapolated sample.
        }
\end{figure}

Working under the manifold hypothesis, a body of work around metric learning \citep{gruffazRiemannianMetricLearning2025a} aims to endow latent spaces with relevant geometric structures. Yet the symmetric nature of these frameworks, \textit{i.e.} that $\text{dist}(a,b)=\text{dist}(b,a)$ and that the underlying trajectory is identical regardless of direction, fails to describe many real world situations. In clinical settings, some pathologies display monotonic evolution, e.g. cognitive scores on Alzheimer's disease. But, when considering the study of possible evolution, there are no guarantees that the extrapolated trajectories will respect this constraint. Clinical markers such as cognitive scores may vary non-monotonically along the computed path, which is violating a known monotonicity constraint. We use this setting to motivate our approach and validate it on controlled datasets exhibiting the same monotonicity structure.

In this work we propose to embed the dynamics of the data directly as a static learned geometry. This geometry needs to take into account asymmetry to inject prior knowledge on the structure about sequential constraints. It ultimately allows longitudinal trajectories to be built from cross-sectional datasets.

To sum up our contributions, we introduce Finsler geometry and Randers metrics as both a principled model and a minimal extension of current frameworks to model asymmetric processes. We then propose a lightweight algorithm to learn latent Randers metrics a posteriori from local scalar ordering. Finally, we evaluate our approach on controlled interpolation and extrapolation settings. We show that if Riemannian geometry can interpolate adequately when endpoints are observed, symmetric geometry cannot reliably extrapolate under directional dynamics. Randers geometry supplies the missing directional inductive bias.

\section{A primer on geometry}
\subsection{Riemann Manifolds}
We consider $\gZ$ a smooth complete sub-manifold of $\R^d$. Hence $ \gZ$ is endowed with a Riemannian metric $G$ that defines a smooth inner product on the tangent space $T_z\gZ$, $\langle\cdot,\cdot\rangle_{z} : (u,v) \mapsto u^TG(z)v $,  where $G(z)$ is a symmetric positive definite matrix for all $z\in \gZ$. We note $\lVert v\rVert_{G(z)} = \sqrt{\langle v,v\rangle_{z}}$. We can consider Riemannian manifolds as curved vector spaces where the energy to cross a certain region is given locally by the metric. We can then define the length $L$ and energy $E$ of a curve $\gamma\in\mathcal{C}^1([0,1],\gZ)$ on the manifold:
\begin{equation}
    \label{def_geo_dst_nrj}
    L(\gamma)=\int_0^1 \lVert \dot{\gamma}_t\rVert_{G(\gamma_t)} dt \quad \text{and} \quad
    E(\gamma)=\int_0^1 \lVert \dot{\gamma}_t\rVert_{G(\gamma_t)}^2 dt \;\text{ ,}
\end{equation}

which in turn allows us to define the geodesic distance between points on the manifold as path of least distance. These minimizers coincide with the critical points of the energy functional for unit speed curves.
For a more in depth introduction, one may refer to \citet{guiguiIntroductionRiemannianGeometry2023}.

\subsection{Finsler Manifolds}
Finsler manifolds are a natural extension of Riemannian ones. They provide a way to smoothly encode asymmetric local distances whilst broadly keeping the same formulation as in Riemannian geometry. One might refer to \citet{ohtaComparisonFinslerGeometry2021} for a more complete 
presentation.

\begin{definition}[Finsler metric]\label{def:finsler_metric}
A function $\FF_z:T_z \gZ\rightarrow \R_+$ at point $z$ is a Finsler metric if: it is $\mathcal{C}^\infty$ on $T\gZ\setminus0$; positive homogeneity : $\FF_z(\lambda v) = \lambda \FF_z(v)$ for $\lambda>0$; and the fundamental tensor $g_{ij}(z,v) = \frac{\partial^2 \FF_z}{\partial v_i\partial v_j}$ is positive definite.
\end{definition}

These metrics induce potentially asymmetric distances. In particular, we do not always have $\FF_z(v) = \FF_z(-v)$ since the homogeneity only applies to positive scalars. Using Euler's theorem on homogeneous function, we have that $\FF^2_z(v)=v^Tg(z,v)v = \lVert v\rVert_{g(z,v)}^2$.

Similarly to Riemannian geometry, shortest paths under the Finsler metric are also minimizers of the energy of the curve under unit speed assumption. 
Consequently, we can define the length and energy of curves and thus geodesic distances on Finsler manifolds by replacing $G(\gamma_t)$ by $g(\gamma_t,\dot{\gamma}_t)$ in Equations \ref{def_geo_dst_nrj}.
This enables to use the same class of algorithm as in the Riemannian case to solve the shortest path problem for marginal additional cost \citep{rygaardGEORCEFastNew2025,arvanitidis2019fast}.

\begin{definition}[Dual metric]\label{def:dual_metric}
The dual metric $\FF^*_z:T^*_z\gZ\rightarrow \R_+$ of a Finsler metric is given by
\begin{equation}
    \forall p\in T^*_z\gZ,\; \FF^*_z(p) = \sup_{\lVert v \rVert \neq 0} \frac{\langle v , p\rangle}{\FF_z(v)}.
\end{equation}
\end{definition}

\begin{definition}[Legendre transform]\label{def:legendre}
The \textit{Legendre transform} $\mathcal{L} : T\gZ \to T^*\gZ$ is uniquely defined by the conditions
\begin{equation}
    \FF_z^*(\mathcal{L}_z (v)) = \FF_z(v), \qquad \langle \mathcal{L}_z(v),v \rangle  = \FF_z^2(v).
\end{equation}
The Legendre transform is a diffeomorphism between the tangent bundle $T\gZ$ and its dual $T^*\gZ$. We can compute it using the fundamental tensor as $\mathcal{L}_z(v) = g(z,v)v$.
In the case where the Finsler metric is Riemannian, the dual metric reduces to the usual co-metric $G^{-1}$ used \textit{e.g.} in Riemannian geodesic shooting \citep{girolamiRiemannManifoldLangevin2011} and $\lVert p\rVert_z^* = \sqrt{p^TG(z)^{-1}p}$.
\end{definition}

\subsection{Randers-Finsler manifolds}
\label{rf_manifold}
\begin{definition}[Randers-Finsler metric]\label{def:randers}
Randers-Finsler metrics are a particular case of Finsler metrics \citep{randers1941asymmetrical}. They are simple extension of Riemannian norm such that, for a Riemannian manifold $(\gZ,G)$, we have:
\begin{equation}
    \FF_{z}(v) = \lVert v\rVert_{G(z)} + \langle\omega(z),v\rangle \,,
\end{equation}
where $\omega:\gZ\rightarrow T_{.}^*\gZ$ is such that $\lVert \omega(z)\rVert_{G^{-1}(z)}<1$ to ensure positivity. These metrics are Finsler metrics. Note that for such metric we have an explicit form of $g$ and $g^{-1}$, enabling efficient implementation. Also the dual metric is still a Randers metric, with explicit and easy to compute expression (\appref{app:prop_randers_metric}).
\end{definition}

With these metrics, the direction only influences the dot product part of the metric. Thus, they can be seen as regular Riemannian norms regularized by a measure of agreement with a low amplitude covector field $\omega$ \citep{chengFinslerGeometry2012}. In particular, we can isolate \textit{sequential hyperplanes} $\mathcal{P}_{z} = \{ v\in T_{z}\gZ \mid  \langle\omega(z),v\rangle=0\}$. Along these hyperplanes, the \textit{1-form} has no influence and the local displacement's energy is only influenced by the Riemannian norm, see Figure \ref{fig:explication_hyperplan}. Therefore, in order to minimize the energy along a trajectory, the tangent vector needs to be on the correct side of $\gP_{z}$ so that $\langle\omega(z),v\rangle \leq 0$, \textit{i.e.} the trajectory needs to be antipodal to $\omega$.

\section{Related work}
\subsection{Metric Learning}
Metric learning is a long-standing field in machine learning \citep{friedman1994flexible,lebanon2002learning,yang2006distance}. It involves finding new data representations that better reflect the task at hand. Although initially focused on linear or at least analytical projections \citep{kulis2013metric}, it has evolved to accommodate the rise of deep neural networks for data projection \citep{kaya2019deep}. However, these methods tend to fail when considering long-range interactions. Their lack of granularity drove the introduction of mixes of different spatially varying metrics to model relevant distances \citep{NIPS2012_ec5aa0b7}. 

Hence, equipped with a Riemannian framework, it becomes possible to describe \textit{local to global} interactions. The difficulty lies in designing a Riemannian metric $G$ that is both computable and expressive enough to characterize the data \citep{gruffazRiemannianMetricLearning2025a}. Building on the data manifold hypothesis \citep{fefferman2016testing}, a body of work aims to pull-back the geometry from the image manifold to the latent space \citep{arvanitidis2018latent,sun2025geometry}. Yet these approaches suffer two drawbacks. First, albeit these metrics are defined on any Riemannian metrics to be pulled back, they often work on the underlying hypothesis that the local Euclidean distance is a correct specification of the image space, while a practical assumption, it is known to be false in general \citep{wang2005euclidean}. Then, from a computational standpoint, they amount to computing the Jacobians of neural networks which scales horrendously poorly with the network size. Consequently other authors have aimed to learn more efficient representation, either by directly working in the latent space \citep{arvanitidis2016locally,kapusniak2024metric,chadebec2022geometric}, learning the metric end-to-end in the latent space \citep{chadebec2022data}, or taking an information geometry point of view \citep{arvanitidis2022pulling}.

\subsection{Finsler Geometry}
This work aims to further the use of Finsler geometry in machine learning. While these tools have received extensive attention from their inception in mathematics \citep{finsler1918ueber,ohtaComparisonFinslerGeometry2021,yeonHistoryBirthFinsler}, they were quickly adopted in general relativity to model spacetime asymmetries \citep{randers1941asymmetrical,pfeifer2019finsler}.
This framework lends itself to model other potentially asymmetric physical phenomenons such as computing the shortest navigation path under low amplitude wind \citep{zermelo1931navigationsproblem,markvorsen2025time}, fire-front propagation \citep{markvorsenFinslerGeodesicSpray2016} or distances on sloped surfaces \citep{matsumoto1989slope}. Such metrics have also been used to model anisotropy in shape analysis and image processing \citep{yang2018geodesic,weber2024finsler,chen2024region}. Recently the links between stochastic Riemannian metric and associated Finslerian geometry have been investigated by \citet{pouplin2022identifying}, along with the associated algorithm to compute geodesics \citep{rygaardGEORCEFastNew2025}. However, learning a Finsler metric from the data has only been scarcely explored, notable examples include: \citet{dages2026harnessing} uses a constant Randers metrics to inject a prior in dimension reduction algorithms ;\citet{zweig2026learning} learns transport maps using a lineage guided tree Finsler metric; and \citet{gahtan2026wildfire} learns the Randers metric associated to fire-front propagation based on wind and topological data.

\section{Methods}
From now on, we will consider a dataset $\gD =\{ (x_i,z_i,s_i)_i\}$ where $x_i\in\gX\subset\R^D$ are the images, $z_i\in\gZ\subset\R^d$ their latent features, and $s_i\in\R$  monotonic scalar values. We also require the $s_i$ to be totally ordered. These values are scores reflecting the sequential dynamic we want to follow in the latent space: time in longitudinal studies, cognitive score for neurological diseases etc. We limit our investigation to phenomena modellable by a monotonic score, that is, for a series of $T$ measurements from a single individual $\pi$, we require $s_\pi^1\le\cdots\le s_\pi^T$.
Being totally ordered is a strong hypothesis. For example, in a clinical setting, it means that we can with certainty compare disease progression between two \textit{different} patients given their scores. We emphasize here that our framework only requires a \textit{cross-sectional dataset} fulfilling the above hypothesis. 

In this section, we first present the underlying Riemannian metrics that will be used. The presented metrics are either extracted or adapted from the Riemannian metric learning literature. We then introduce our core contribution: a learning algorithm for the \textit{1-form} $\omega$ of the Randers metric following the insights from Figure \ref{fig:explication_hyperplan}.

\subsection{The Riemannian metrics}
\label{section:learning}
We now assume that the latent space $(\gZ,G)$ is a Riemannian manifold with metric $G$. Usually, the implicit Euclidean assumption considers that we have $G=I_d$ and thus that geodesics are straight lines. Riemannian metric learning aims to learn appropriate $G$. 

We consider a VAE \citep{kingma2013auto} $\gX \xrightarrow[]{E}\gZ\xrightarrow[]{D}\gX$, where $E$ is the encoder such that $E(x_k) =(z_k,\Sigma_k) $ with $\Sigma_k$ the latent covariance matrix associated to the features $z_k$. It allows to define various metric on the latent space. We first use the diagonal metric from \citet{arvanitidis2016locally}(\textbf{LAND}):
\begin{equation}
    \label{eq:LAND_metric}
    G^{-1}_{ii}(z) = \sum_k \exp\left(-\frac{\lVert z - z_k\rVert_2^2}{2\tau^2_k}\right) ( z_k^{(i)} - z^{(i)})^2 + \lambda\;\text{.}
\end{equation}
It enforces the data manifold support using Gaussian-weighted centroids. We also consider the variants from \citet{kapusniak2024metric}(\textbf{LAND RBF}), where the interpolation weights are learned, and from \citet{chadebec2022geometric}(\textbf{CovCentroids}) that uses the learned covariance matrix $\Sigma_k$ instead of the local covariance of the latent samples. The parameters $\tau_k$ control each centroids' importance while $\lambda$ is a regularization parameter\footnote{For larger dataset, the sum can be computed on a subset of points using \textit{K-Medoids}}. Also, given the Jacobian of the decoder $J_f(z) = \frac{\partial f}{\partial z}$, we have that $G_{\text{pb}}(z)=J_f(z)J_f(z)^T$ defines a Riemannian metric. To be used in practice, these pullback-based metrics need specific calibration \citet{arvanitidis2018latent,arvanitidis2022pulling,sun2025geometry}. Still, they require to differentiate through potentially large networks. In our setup, these metrics don't fit on our available compute memory to compute geodesics.

To alleviate the significant cost of recomputing the Jacobian matrices at each point we consider a computationally tractable variant that we subsequently call the \textbf{PBCentroids} metric:
\begin{equation}
\label{eq:metric_form_pullback}
    G^{-1}(z) = \sum_k\exp{\left( -\frac{\lVert z-z_k\Vert^2}{2\tau^2_k} \right)} G_{\text{pb}}(z_k)^{-1} + \lambda I_d\,.
\end{equation}

We provide a more detailed overview of these methods and their parameters in \appref{app:metrics_description}.

\subsection{The Randers metric}
\label{section:randers_learning}
For ease of training, we parametrize $\omega$ by decomposing the direction and amplitude such that
\begin{equation}
\label{eq:omega_polar_form}
    \omega_\theta(z) = \rho_\theta(z) \phi_\theta(z)\,.
\end{equation}
This factorization decorrelates the sequential orientation from its strength. While $\phi_\theta:\gZ\rightarrow T_z\gZ$ is meant to showcase the sequential dynamic. $\rho_\theta:\gZ\rightarrow [0,1]$ aims to reflect how locally orientable the space is, effectively weighting the influence of the \textit{1-form} in the metric $\FF$. To enforce that the learned metric is positive, we learn $\phi_\theta$ as:
\begin{equation}
    \phi_\theta(z) = \frac{1}{\lVert \tilde{\phi}_\theta(z)\rVert_{G^{-1}(z)}+\varepsilon}\tilde{\phi}_\theta(z) \,\text{ ,}
\end{equation}
where $\varepsilon<<1$ and $\tilde{\phi}_\theta$ is a neural network such that we enforce $\lVert \omega_\theta(z)\rVert_{G^{-1}(z)}<1$.

Following the discussion of \ref{rf_manifold}, we want to learn locally separating hyperplanes $\gP_{z}$ for $z\in\gZ$. Inspired by the SVM literature \citep{hearst1998support,xu2012distance}, we thus learn $\tilde{\phi}_\theta$ through a local hinge loss such that :
\begin{equation}
    l_{\tilde{\phi}}(z_i)= \frac{1}{K_\phi}\sum_{z_j\in\text{KNN}(z_i)} \max(0, 1 - y_i^j\times\tilde{\phi}_\theta^\top(z_i) G(z_i)(z_j-z_i))\;,
\end{equation}
where $y_i^j =\sign (s_i-s_j) \in\{\pm1\}$. The loss is only computed on the $K_\phi$ closest neighbors of each point to enforce locality and use the Riemannian dot product.

This results in a vector field $\phi_\theta$ on the manifold that is able to locally discriminate the "before" from the "after" effectively orienting the latent space. Yet usually not all the latent space can be correctly orientated. To account for this we regularize the strength of the asymmetry of the metric through $\rho_\theta$ by regressing the local accuracy of the separating hyperplane. That is, at point $z_i$ we compute the local accuracy of the hyperplane and we consider the region orientable if it is above $0.5$. We thus learn to regress its significant portion mapped to $[0,1]$ :

\begin{equation}
    l_\rho(z_i)  =\left\lVert \rho_\theta(z_i) - 2\times\max\left(0,\text{ACC}_{\tilde{\phi}}(z_i)-\frac{1}{2}\right) \right\lVert^2\,
\end{equation}
\begin{equation}
    \text{where   } \quad \text{ACC}_{\tilde{\phi}(z_i)}=\frac{1}{K_\phi}\# \left\{ y_i^j\times\tilde{\phi}_\theta(z_i)^\top G(z_i)(z_j-z_i) > 0 \mid z_j\in\text{KNN}(z_i)\right\} .
\end{equation}

Consequently, when $\phi_\theta$ produces a good separating hyperplane, that means that the latent space is locally orientable and we will have $\rho_\theta \approx 1$. On the contrary, when the sequential dynamic is not structuring enough, we will have $\rho_\theta\approx 0$, effectively falling back to the regular Riemannian geometry.

We first learn $\phi_\theta$ by optimizing $l_\phi$ then we learn $\rho_\theta$ by optimizing $l_\rho$. This yields in practice a lightweight algorithm as it essentially amounts to learning two small networks $\phi$ and $\rho$ in small dimensions with possibly pre-extracted latent features. We investigate the influence of the polar decomposition in \appref{app:ablations}.

\subsection{Interpolation}
\label{section:interpolation}
To assess the validity of our framework, we quantify the regression of the sequential factors along interpolated trajectories.
So we sample $z_0$ and $z_1$ such that $s_0<s_1$. We then compute the latent geodesic $\gamma_t$ from $z_0$ to $z_1$ using $T=25$ time points. The geodesics are initialized by the shortest path of the data's KNN-graph \citep{kimDeepGenerativeGeodesics2024a}. They are then refined using the GEORCE solver from \citet{rygaardGEORCEFastNew2025} for diagonal metrics, or by minimizing energy of the curve under BSpline parametrization for dense metrics \citep{torchcurves}. Computation details are available in \appref{app:geodesic_comp}.

\begin{restatable}[Path invariance]{proposition}{pathinvariance}\label{prop:path_invariance}
Let $U$ be a simply connected domain of $\gZ$. If there exists a smooth potential function $\psi$ such that $\omega_\theta = \nabla\psi$, then the Randers and Riemannian geodesics between fixed endpoints coincide up to reparametrization. In particular $dst_\FF(z_0,z_1) = dst_G(z_0,z_1) + \psi(z_1) -\psi(z_0)$.
\end{restatable}

This result, proved in \appref{app:proof_prop}, demonstrates that if the learned vector field $\omega_\theta$ approximates the gradient of the sequential factors $s$, ie $\omega\approx\nabla s$, then the interpolated Randers trajectory will be identical to the Riemannian ones.

\subsection{Extrapolation}
The learned vector field enables the computation of dynamics-aware extrapolation. Here we use geodesic guided shooting techniques. Indeed, given the dual metric $\FF_*$, we can show that, for the Hamiltonian $H(p,q) = \frac{1}{2} \FF_*(q,p)^2$ with $(q,p)\in T^*\gZ$, the flow 
\begin{equation}
    \label{eq:hamiltonian_system}
    \dot{p}_t = -\frac{\partial H}{\partial q}(p_t,q_t) \quad \; \dot{q}_t = \frac{\partial H}{\partial p}(p_t,q_t)
\end{equation}

yields geodesic trajectories \citep[section 2.3]{hryniewicz2013global}. As the functional is not separable, we integrate the system using the \textit{generalized leapfrog algorithm} with implicit steps \citep{girolamiRiemannManifoldLangevin2011,chadebec2022data,cobbIntroducingExplicitSymplectic2019}.

We can initialize this flow using the learned dynamic encoded by $\omega$. So if we note $\tilde{\omega}_0 = \frac{\omega(q_0)}{\lVert \omega(q_0) \rVert_2}$, the initial speed  $\widetilde{v}_0 \in T_{q_0}\gZ$ is sampled such that $\widetilde{v}_0 \mid q_0 \sim \text{vMF}(-\tilde{\omega}_0, \kappa)$ follows a von Mises Fisher distribution \citep{pinzonFastPythonSampler2023} centered around the direction of evolution given by $-\tilde{\omega}_0$ and of concentration $\kappa$. We essentially sample $\tilde{v}_0$ in the cone of direction $-\tilde{\omega}_0$ and aperture $\kappa$. We then rescale the initial speed to have a norm $\alpha$, $v_0 = \alpha \tilde{v}_0$. This initial speed $v_0\in T_{q_0}\gZ$ is then pushed to the dual tangent space using the Legendre transform $p_0 = g(q_0,v_0)v_0\in T_{q_0}^*\gZ$. 

We also choose $\kappa$ to match a target concentration $\eta$ that verifies $\eta = \mathbb{E}_{v\sim \text{vMF}(-\tilde{\omega}_0,\kappa)}[\text{cosim}(-\tilde{\omega}_0,v)]$. The parameter $\eta$ allows to tune the angle between the sampled direction of evolution $-\tilde{\omega}_0$ and the initial speed $v_0$. A value of $\eta=1$ means that the sampled direction is always aligned with $-\tilde{\omega}_0$, while $\eta=0$ means that the sampled direction is uniformly distributed on the sphere. The following proposition proved in \appref{app:kappa_vmf_extrapolation} allows to find the expression of $\kappa$ given $\eta$.

\begin{proposition}[Concentration parameter]\label{prop:kappa_eta}
For a given concentration parameter $\eta$, we have that
\begin{equation}
    \eta = I_{\frac{d}{2}}(\kappa) / I_{\frac{d}{2}-1}(\kappa), \quad \text{such that} \quad \kappa \approx \frac{\eta d - \eta^3}{1-\eta^2},
\end{equation}
where $I_d$ is the modified Bessel function of the first kind of order $d$.
\end{proposition}

To ensure fair comparison with Riemannian settings, we follow the usual setup and sample $\widetilde{v}_0 \mid q_0 \sim \mathcal{N}(0, G^{-1}(q_0))$. We also rescale the initial speed so that $v_0 = \alpha \frac{\widetilde{v}_0}{\lVert\widetilde{v}_0\rVert_2}$ which yields $p_0 = G(q_0)v_0$.

In order to mitigate the effects of numerical approximation, we also resample the speed after each integration step with momentum $\zeta$, so that at time step $t$, we have :
\begin{equation}
    v_{t+1} = \zeta v_t + (1-\zeta) v_{\text{sampled}}.
\end{equation}

\subsection{Evaluation setup}
\label{section:eval_setup}
Given a trajectory in the latent space, either from interpolation or extrapolation, it is first decoded back to the image space. Then we can recover the unknown sequential factors $s_t^\gamma$.
We evaluate the discrete trajectories of sequential factors $s^\gamma= (s^\gamma_k)_{k\in\{0,\cdots,T-1\}}$ using the monotonicity score $\tau(\gamma)$ defined as the Kendall rank correlation coefficient
between $s^\gamma$ and $\{0,\cdots,T-1\}$. In the extrapolation setup, we also report the relative variation of the sequential factors as $100\times \frac{s^\gamma_T - s^\gamma_0}{s^\gamma_0}$. This allows to monitor the ability of the extrapolated trajectory to actually produce meaningful diverse trajectories. We also report the Fréchet Wavelet Distance (FWD) \citep{veeramacheneni2023fr} to ensure that the generated images are within the data distribution. A summary of the learning algorithms and evaluation procedure is available in \appref{app:algorithms}.

\section{Experiments}

\begin{figure}[ht]
     \centering
        \includegraphics[width=\textwidth]{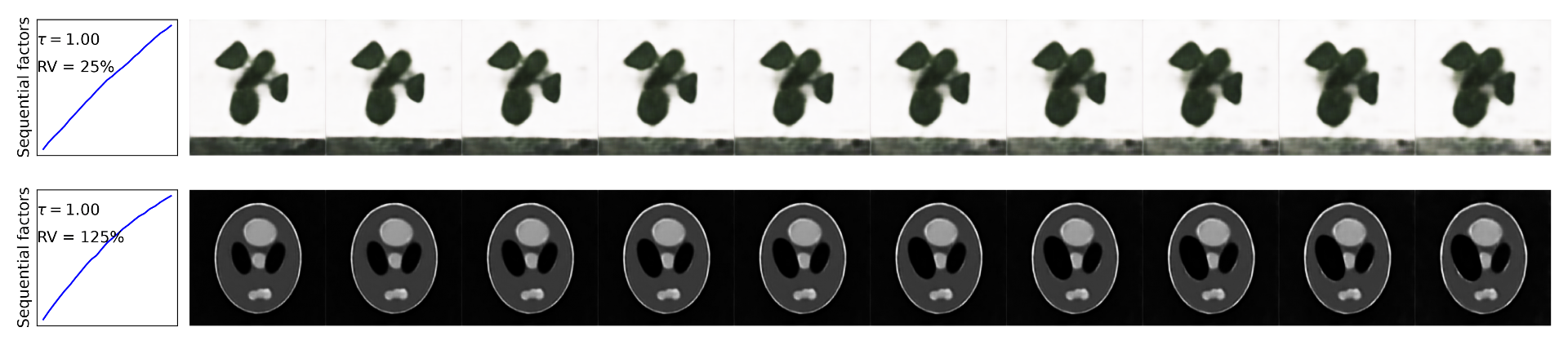}
        \caption{Example of extrapolated trajectory on the KOMATSUNA dataset (\textit{top}) and Shepp-Logan (\textit{bottom}) using the asymmetric framework with \textit{LAND} base metric. Left columns displays the sequential factors along the trajectory where $\tau$ is the Kendall Tau correlation and \textit{RV} is the relative variation.}
        \label{fig:example_extrapolation}
\end{figure}

As the latent space is structured by the metric $G$ and the sequential dynamics are described by $\omega$, we can provide dynamics-aware possible evolutions given a single acquisition point (Figure \ref{fig:shepplogan_inter_then_extra}), or even create synthetic longitudinal cohorts given a cross-sectional cohort (Figure \ref{fig:example_extrapolation}).

Following other work in the metric learning literature 
\cite{diepeveen2025score,arvanitidis2018latent,sun2025geometry}, we first focus on a controlled synthetic setting, here fulfilling the framework’s hypothesis : monotonic dynamics, totally ordered and continuous sequential factor easily retrievable from the image and no label noise. This setting and the provided experiments allow isolating the causes of failure of usual approaches whilst showcasing the desired behavior achieved by the introduced Randers metric. The framework is motivated by clinical progression problems; our experiments establish feasibility in controlled medical-imaging-like and biological-growth settings.

We experiment using the 5 metrics described in Section \ref{section:learning} using a VAE trained from scratch. Architecture and training details are available in \appref{app:experimental_details}. For each metric we report the results using the base metric and its Finsler variant with the learned vector field $\omega$. We also compare our approach to other baselines : \textit{Wasserstein gradient flow learning} \citep{huizing2026stories} and \textit{HyperSphericalVAE} \citep{sablica2025hyperspherical}, see details at \appref{app:wasserstein}. We ablate the hyper-parameters influence and noise robustness in \appref{app:ablations}. 

\subsection{Shepp-Logan dataset}
To simulate a medical imaging dataset fitting our criterion we use variations of the Shepp-Logan phantom \citep{shepp1974fourier}. It consists of a model of a human head with the skull, the brain, and several ellipses representing different brain sections. We generate images ($n_\text{train} = 1000$ and $n_\text{test} = 500$) by varying the colour and shape of the ellipses and adding random Gaussian noise. The sequential factors $(s_i)$ are here the area of the top left black ellipsis. They aim to represent tumour size.

\subsection{KOMATSUNA dataset}
The KOMATSUNA dataset \citep{uchiyama2017easy} consists of a series of images ($n_\text{train} = 720 $ and $n_\text{test} = 180$) of the growth process of the komatsuna plant. We use as sequential factors $s_i$ the leaf area of the plant as it fulfills the framework's hypothesis.

\subsection{Results}
{%
\setcitestyle{numbers,square,citesep={,}}
\begin{table}[ht]
\centering
\caption{Combined extrapolation and interpolation results on the Shepp-Logan dataset. Here, $d=64, \alpha=5.0, \eta=0.999$, and $\zeta=0.9$.}
\label{tab:combined_results}
\begin{tabular}{lccccc}
\toprule
    & & \multicolumn{2}{c}{Interpolation} & \multicolumn{2}{c}{Extrapolation} \\
    \cmidrule(lr){3-4} \cmidrule(lr){5-6}
    Base metric & Finsler & Kendall Tau ($\uparrow$) & FWD ($\downarrow$) & Kendall Tau ($\uparrow$) & Relative Variation (\%) \\
    \midrule
    \multirow{2}{*}{Euclidean} & \xmark & $0.94 \pm 0.19$ & $0.35$ & $0.02 \pm 0.81$ & $0.09 \pm 5.72$ \\
     & \checkmark & $1.0 \pm 0.02$ & $0.41$ & $1.0 \pm 0.02$ & $85.59 \pm 33.52$ \\
\midrule
    \multirow{2}{*}{LAND \citep{arvanitidis2016locally}} & \xmark & $0.46 \pm 0.29$ & $0.23$ & $-0.04 \pm 0.8$ & $-0.14 \pm 6.04$ \\
     & \checkmark & $0.46 \pm 0.29$ & $0.19$ & $1.0 \pm 0.0$ & $90.00 \pm 44.26$ \\
\midrule
    \multirow{2}{*}{LAND RBF \citep{kapusniak2024metric}} & \xmark & $0.46 \pm 0.29$ & $0.21$ & $0.0 \pm 0.8$ & $-0.11 \pm 5.59$ \\
     & \checkmark & $0.46 \pm 0.3$ & $0.22$ & $1.0 \pm 0.02$ & $83.47 \pm 30.76$ \\
\midrule
    \multirow{2}{*}{CovCentroids \citep{chadebec2022geometric}} & \xmark & $0.46 \pm 0.29$ & $0.17$ & $0.03 \pm 0.79$ & $0.15 \pm 6.68$ \\
     & \checkmark & $0.45 \pm 0.29$ & $0.2$ & $1.0 \pm 0.03$ & $74.53 \pm 74.71$ \\
\midrule
    \multirow{2}{*}{PBCentroids} & \xmark & $0.96 \pm 0.09$ & $0.21$ & $0.01 \pm 0.79$ & $0.32 \pm 4.55$ \\
     & \checkmark & $0.77 \pm 0.31$ & $0.23$ & $0.59 \pm 0.55$ & $28.75 \pm 70.48$ \\
\midrule
    \multirow{1}{*}{Wasserstein \citep{huizing2026stories}} & \xmark & $\cdots$ & $\cdots$ & $0.97 \pm 0.15$ & $25.54 \pm 24.04$ \\
\midrule
    \multirow{1}{*}{HyperSpherical \citep{sablica2025hyperspherical}} & \xmark & $1.0 \pm 0.04$ & $0.35$ & $-0.03 \pm 0.75$ & $-0.48 \pm 53.77$ \\
\bottomrule
\end{tabular}
\end{table}
}%

{%
\setcitestyle{numbers,square,citesep={,}}
\begin{table}[ht]
\centering
\caption{Combined extrapolation and interpolation results on the KOMATSUNA dataset. Here, $d=64, \alpha=5.0, \eta=0.999$, and $\zeta=0.9$.}
\label{tab:combined_results_komatsuna}
\begin{tabular}{lccccc}
\toprule
    & & \multicolumn{2}{c}{Interpolation} & \multicolumn{2}{c}{Extrapolation} \\
    \cmidrule(lr){3-4} \cmidrule(lr){5-6}
    Base metric & Finsler & Kendall Tau ($\uparrow$) & FWD ($\downarrow$) & Kendall Tau ($\uparrow$) & Relative Variation (\%) \\
    \midrule
    \multirow{2}{*}{Euclidean} & \xmark & $0.98 \pm 0.06$ & $4.67$ & $-0.05 \pm 0.78$ & $0.16 \pm 8.63$ \\
     & \checkmark & $0.98 \pm 0.06$ & $4.71$ & $1.0 \pm 0.05$ & $138.56 \pm 130.66$ \\
\midrule
    \multirow{2}{*}{LAND \citep{arvanitidis2016locally}} & \xmark & $0.76 \pm 0.26$ & $4.02$ & $-0.11 \pm 0.78$ & $-1.01 \pm 8.01$ \\
     & \checkmark & $0.57 \pm 0.29$ & $4.29$ & $1.0 \pm 0.01$ & $112.84 \pm 97.52$ \\
\midrule
    \multirow{2}{*}{LAND RBF \citep{kapusniak2024metric}} & \xmark & $0.98 \pm 0.04$ & $4.28$ & $-0.12 \pm 0.76$ & $-0.36 \pm 7.52$ \\
     & \checkmark & $0.99 \pm 0.03$ & $4.53$ & $1.0 \pm 0.01$ & $141.37 \pm 140.31$ \\
\midrule
    \multirow{2}{*}{CovCentroids \citep{chadebec2022geometric}} & \xmark & $0.42 \pm 0.33$ & $3.76$ & $-0.05 \pm 0.78$ & $-0.02 \pm 10.08$ \\
     & \checkmark & $0.43 \pm 0.41$ & $3.47$ & $0.99 \pm 0.13$ & $332.15 \pm 584.35$ \\
\midrule
    \multirow{2}{*}{PBCentroids} & \xmark & $0.44 \pm 0.38$ & $6.07$ & $-0.06 \pm 0.76$ & $1.53 \pm 18.50$ \\
     & \checkmark & $0.48 \pm 0.38$ & $5.65$ & $0.36 \pm 0.58$ & $63.83 \pm 192.03$ \\
\midrule
    \multirow{1}{*}{Wasserstein \citep{huizing2026stories}} & \xmark & $\cdots$ & $\cdots$ & $0.87 \pm 0.37$ & $351.62 \pm 489.63$ \\
\midrule
    \multirow{1}{*}{HyperSpherical \citep{sablica2025hyperspherical}} & \xmark & $0.99 \pm 0.03$ & $7.63$ & $0.18 \pm 0.72$ & $98.25 \pm 170.43$ \\
\bottomrule
\end{tabular}
\end{table}
}%
The results are displayed in \tabref{tab:combined_results} for the Shepp-Logan dataset and in \tabref{tab:combined_results_komatsuna} for the KOMATSUNA dataset. We can draw similar conclusions for both datasets. The naive symmetric \textit{Euclidean} framework reaches top performances in interpolation. This is expected as the Gaussian prior on the latent space tends to convexify the latent space, which mitigates data manifold support issues. Furthermore, VAE are known to have relatively disentangled representations \citep{burgess2018understanding}, hence the good interpolation performances. In most settings the addition of the asymmetry doesn't change the interpolation results. This is expected per \cref{prop:path_invariance}. Indeed, if we can assume $\omega\approx\nabla s$, then the Finslerian and Riemannian interpolated trajectories are the same, which yields the observed identical scores. The discrepancies observed on \textit{PBCentroids} in \tabref{tab:combined_results} and \textit{LAND} in \tabref{tab:combined_results_komatsuna} are to be attributed to the base metric. Indeed the learning of $\omega_\theta$ is induced by the angles defined by $\langle\cdot,\cdot\rangle_G$. A mis-specification of the metric for a given dataset will yield a spurious learning of the vector field.

The proposed method proves to be more useful in the extrapolation regime. The results demonstrate that the asymmetric frameworks yield the correct expected behavior. Indeed, the symmetric shooting methods actually amount to a non-homogeneous random walk on the manifold, thus on average describing a diffusion process. Incorporating prior information in the geometry allows to provide sequentially consistent trajectories. The results hold even in the presence of moderate noise on the labels, see details in \appref{app:ablations}.

The directional component dominates the choice of base metric in the tested regimes, whereas interpolation remains sensitive to the base geometry. Apart for the \textit{PBCentroids} one, they all achieve high monotonicity score whilst exploring a wide range of extrapolated data as shown by the high and diverse relative variation score.
The hyper-parameter search in \appref{app:ablations} reveals that the tuning of the concentration parameter $\kappa$ is crucial to get good performances. For example, on the Shepp-Logan dataset the relative variation drop from $74.53\pm74.71\%$ to $37.48\pm12.45\%$ on the \textit{CovCentroids} geometry when $\eta$ changes from $0.999$ to $0.5$ while still maintaining high Kendall Tau.
It also demonstrates that, unsurprisingly, trajectories reaching further, or equivalently with higher speed magnitude, achieve broader exploration of the space whilst respecting the learned dynamic : the relative variation drops from $74.53\pm74.71\%$ to $18.88\pm5.63\%$ on the \textit{CovCentroids} geometry when $\alpha$ changes from $5.0$ to $1.0$, while still maintaining high Kendall Tau again.

\section{Limitations and Future work}

We see two important lines of future work. First we need to assess the scalability of the presented method to bigger datasets. To open up more potential datasets, we need to investigate how to relax the total ordering hypothesis on the scalar values $s_i$. So far it is only used to compute local pseudo-labels in the learning of $\omega$. One could imagine computing the loss only on a single patient's samples. But this would require very important sampling frequency so as to maintain the desired locality. We could extend this framework to work with binned discrete clinical scores, such as LI-RADS or MMSE, as they enable inter-patient comparison. The problem lies in having adequate evaluation since these quantities are hard to retrieve from images \citep{wang2023fully} and are subject to noisy labels \citep{low2025lr}, although we hypothesize that the demonstrated robustness to noise could alleviate such noisy-label issues.

\section{Conclusion and discussion}

Faithful latent extrapolation under sequential dynamics constraints requires modelling directionality beyond curvature. We embed such dynamics as a static learned asymmetric geometry through Finsler-Randers metrics, allowing us to model asymmetric transition cost whilst avoiding the need for longitudinal datasets. This lightweight directional regularization outperforms symmetric frameworks in our experiments. It suggests that asymmetry is an important ingredient in modelling progression. This opens the door to richer geometric priors for trajectory extrapolation in clinical settings.

\subsection*{AI use statement}



In this work, we used generative AI tools to provide feedback on research.
We have not used generative AI tools to develop the conceptual framework developed in the paper, and the rest of the required disclosure tasks are not applicable to this work.
Additionally, we used generative AI tools to edit software code, summarize existing literature, edit the research paper to improve readability, identify relevant literature, format references. We have reviewed all AI-assisted work. We take responsibility for the final content of this work,
including text, claims or artifacts produced with the aid of generative AI.

\subsection*{Reproducibility statement}
\label{app:code}

The code is available at \url{https://github.com/Theaublanchard/finsler_learning}. The main algorithms are included as pseudo-code in \appref{app:algorithms}. 


\bibliography{iclr2027_conference}
\bibliographystyle{iclr2027_conference}
\newpage

\appendix

\etocsetlocaltop.toc{part}          
\etocdoesnotcheckemptiness
\etocsettocstyle{\section*{Table of Contents (Appendix)}}{}
\etocsettocdepth{section}
\localtableofcontents               

\section{Properties of Randers metric}
\label{app:prop_randers_metric}
\subsection{Randers-Finsler metrics are Finsler metrics}
\label{app:rf_are_f}
The proof is adapted from \citet{weber2024finsler}.
We aim to show that Randers metrics are Finsler metrics. Triangle inequality and positive homogeneity are trivial. Let's prove the positivity.

Let $(z,u)\in T\gZ$. We note $v = G(z)u$. Then :
\begin{align}
    \FF(z,u) &= \FF(z,G(z)^{-1}v) \\
    &= \lVert G(z)^{-1}v \rVert_{G(z)} + \langle \omega(z),G(z)^{-1}v\rangle \label{app:pos_rf}\\
    &= \sqrt{v^TG(z)^{-1}v} + \omega(z)^TG(z)^{-1}v \\
    &= \lVert G(z)^{-\frac{1}{2}}v\rVert + \langle G(z)^{-\frac{1}{2}}\omega(z), G(z)^{-\frac{1}{2}}v\rangle\\
\end{align}
Using the Cauchy-Schwartz inequality we get :
\begin{equation}
    |\langle \omega(z),G(z)^{-1}v\rangle | \leq \lVert G(z)^{-\frac{1}{2}}\omega(z)\rVert\, \lVert G(z)^{-\frac{1}{2}}v\rVert
\end{equation}
Because by hypothesis we have $\lVert\omega(z)\rVert_{G(z)^{-1}} < 1$, we have $\lVert G(z)^{-\frac{1}{2}}\omega(z)\rVert<1$. Thus :
\begin{equation}
   \mid \langle G(z)^{-\frac{1}{2}}\omega(z), G(z)^{-\frac{1}{2}}v\rangle \mid =  \mid \langle \omega(z),G(z)^{-1}v\rangle \mid \leq \lVert G(z)^{-\frac{1}{2}}v\rVert
\end{equation}
Thus, if $u\neq 0$, we get $\FF(z,u) >0$. So finally Randers metrics are Finsler metrics.

\subsection{Fundamental tensor of a Randers metric}
\label{app:fund_tensor}
For clarity we use the same notation as in \citet{ohtaComparisonFinslerGeometry2021}. That is, we assume a Randers-Finsler metric in the class of $(\alpha,\beta)$-metric where $\alpha_{z}(u,v)=a_{ij}(z)u^iv^j$ is a Riemannian metric and $\beta(z,v)=b_i(z)v^i$ is a \textit{1-form} such that $\FF(z,v) = \sqrt{\alpha_{z}(v,v)} + \beta(x,v)$.
If we note $A = (a_{ij})$ and $b = (b_i)$, we have $\FF(z,v)=\sqrt{v^TA(z)v} + b(z)^Tv$.

 Then, following the work of \citet{baoIntroductionRiemannFinslerGeometry2000}, if we note :
\begin{equation}
    \tilde{l}_i = \frac{a_{ij}v^j}{\lVert v\rVert_\alpha}\, \quad \text{and} \quad     l_i = \tilde{l}_i + b_i\,.
\end{equation}

Then we have :
\begin{equation}
    g_{ij} = \frac{\FF(z,v)}{\lVert v\rVert_\alpha} (a_{ij} - \tilde{l}_i\tilde{l}_j) + l_i l_j.
\end{equation}

And if we note :
\begin{equation}
    (a^{ij}) = (a_{ij})^{-1} \; ; \; b^i = \sum_{j=1}^n a^{ij}b_j \; ; \; \lVert \beta \rVert_\alpha = \sqrt{\sum_{i=1}^nb^ib_i}
\end{equation}
We have:
\begin{multline}
    g^{ij} = \frac{\lVert v\rVert_\alpha}{\FF(z,v)}a^{ij}(x) + \frac{\beta(v) + \lVert\beta\rVert_\alpha^2(z) \lVert v\rVert_\alpha}{\FF^3(z,v)}v^iv^j-
    \frac{\lVert v\rVert_\alpha}{\FF^2(z,v)}(b^i(z)v^j + b^j(z)v^i)\\
\end{multline}

\subsection{Dual Randers Metric}
\label{app:dual_randers_metric}
The dual metric $\FF_*$ of a Finsler metric is defined as :
\begin{equation}
    \forall (q,p)\in T_*\gZ,\quad \FF_q^*(p) = \sup_{\lVert u \rVert \neq 0} \frac{\langle u , p\rangle}{\FF_q(u)}.
\end{equation}

Fortunately, the dual metric of a Randers metric is also a Randers metric  \citep{yang2018geodesic}. More precisely :
\begin{equation}
     \forall (q,p)\in T_*\gZ,\quad \FF_q^*(p) = \lVert p \rVert_{G_*(q)} + \langle \omega_*(q) , p \rangle.
\end{equation}

Where we have :
\begin{equation}
    \delta = 1 - \langle \omega, G^{-1} \omega \rangle \;, G_* = \frac{(G^{-1}\omega)(G^{-1}\omega)^\top + \delta G^{-1}}{\delta^2}\; , \omega_* = -\frac{G_*^{-1}G^{-1}\omega}{\delta}.
\end{equation}

The dual metric depends on the amplitude of the vector field $\lVert\omega(x)\rVert_{G^{-1}(x)}$. Consequently, to reduce numerical instabilities in the extrapolation settings where we use the dual vector field $\omega_*$, we limit the vector field amplitude by capping the value of $\lVert\omega(x)\rVert_{G^{-1}(x)}$ to $0.9$.

\section{Path invariance}
\label{app:proof_prop}
\pathinvariance*

\begin{proof}
Let $z_0,z_1\in U \subset \gZ$. Consider a curve $\gamma : [0,1] \to\gZ$ such that $\gamma(0)=z_0$ and $\gamma(1)=z_1$. Then if we suppose that $\omega=d\psi$:
\begin{align*}
    L_\FF(\gamma) &= \int_0^1 \FF_{\gamma_t}(\dot{\gamma}_t) dt\\
    &=\int_0^1 \sqrt{\dot{\gamma}_t^\top G(\gamma_t) \dot{\gamma}_t} + \langle \omega(\gamma_t),\dot{\gamma}_t\rangle dt\\
    &=L_G(\gamma) +  \int_0^1 \langle\nabla\psi(\gamma_t),\dot{\gamma}_t\rangle dt \\
    &=L_G(\gamma) + \int_0^1 \frac{d}{dt}\psi(\gamma_t)dt\\
    &=L_G(\gamma) + \psi(z_1)-\psi(z_0).
\end{align*}

Consequently, minimizing $L_\FF$ is equivalent to minimizing $L_G$. The Randers part of the metric only influences the distance not the trajectory.
\end{proof}
\section{Description of the metrics used}
\label{app:metrics_description}

\subsection{Metrics expression}
We here present the different metrics used to compare our method. We first consider the setup of a deterministic generator. In this setup we consider a VAE $\mathcal{X} \xrightarrow{g}\gZ\xrightarrow{f}$, where for a point $x\in\mathcal{X}$, We have $g(x) = (z, \Sigma^{-1})$ and $\hat{x} = f(z + \varepsilon \Sigma^{-1})$ with $\varepsilon \sim \mathcal{N}(0, I)$.

\paragraph{LAND}
The diagonal elements of the LAND metric of \citet{arvanitidis2016locally} are defined as :
\begin{equation}
    G^{-1}_{ii}(z) = \sum_k \exp\left(-\frac{\lVert z - z_k\rVert_2^2}{2\tau^2_k}\right) ( z_k^{(i)} - z^{(i)})^2 + \lambda\;\text{.}
\end{equation}
Here $z^{(i)}$ denotes the i-th coordinates of the point $z$, and $z_k^{(i)}$ the i-th coordinates of the k-th centroid.

\paragraph{LAND RBF} This allows us to introduce the LANDRBF metric of \citet{kapusniak2024metric} :
\begin{equation}
    G^{-1}(z) = \text{diag}(h(z)) + \lambda I\,,
\end{equation}
where :
\begin{equation}
    h(z) = \sum_k w_k \exp\left(-\frac{\lVert z - z_k\rVert_2^2}{2\tau_k^2}\right)\,.
\end{equation}
The positive weights are then optimized by minimizing the following loss function :
\begin{equation}
    L(\{w_k\}) = \sum_{z_i\in\gZ} (h(z_i) - 1)^2\,.
\end{equation}

\paragraph{CovCentroids} The Covariance Centroids metric from \citet{chadebec2022geometric} replaces the local covariance of the LAND metric by the learned covariances of the encoder : 
\begin{equation}
    G^{-1}(z) = \sum_k \exp\left(-\frac{\lVert z - z_k\rVert_2^2}{2\tau^2_k}\right) \Sigma_k^{-1} + \lambda I
\end{equation}

\paragraph{PBCentroids} We can pullback the geometry of the image space$\mathcal{X}$ to the latent space $\gZ$ using the pullback of the metric tensor. Under the hypothesis that the decoder is a local diffeomorphism and that the geometry of the image space is Euclidean, we can then define the pullback metric tensor as follows:
\begin{equation}
    G_{\text{pb}}(z) = J_f(z)^\top J_f(z)
\end{equation}
where $J_f(z)$ is the Jacobian of the decoder at point $z$. As this is expensive to compute and ill-calibrated to support the data-manifold, we can instead compute the pullback metric tensor at centroids and then interpolate it in the latent space. The cometric is then defined as follows:
\begin{equation}
    G^{-1}(z) = \sum_k \exp\left(-\frac{\lVert z - z_k\rVert_2^2}{2\tau^2_k}\right) G^{-1}_{\text{pb}}(z_k) + \lambda I
\end{equation}

For the interested reader, we also provide a broader overview including pullback-based metrics. Whilst these metrics have favorable theoretical properties, they require loads of compute. 
\paragraph{WarpedPB} The centroid approach is ad-hoc and relies on correct tuning of the temperature parameters $\tau_k$. Instead, it has been proposed by \citet{sun2025geometry} to add an additional dimension to the decoder so that we now have $\tilde{f} = (f, s)$. It verifies $s(z) \approx 0$ for $z$ in the data-manifold and $s(z) \gg 0$ for $z$ off the data-manifold. We can then define a cometric as follows:
\begin{equation}
    G(z) = J_{\tilde{f}}(z)^\top J_{\tilde{f}}(z) + \lambda I
\end{equation}

The question is then to correctly design and learn such function $s$. The authors propose two approaches. First they propose to train a GAN from the 
latent space to the data space and use the learned discriminator as a function $s$. The second approach is to use a Gaussian Process. We thus obtain $s$ as the variance of the GP at point $z$. The GP here uses a RBF kernel with a temperature parameter $\tau$ such that :
\begin{equation}
    s(z) = K(z,z) - K(z,Z)\left[K(Z,Z) + \tau I\right]^{-1}K(Z,z)
\end{equation}
where $Z = [z_1,\cdots, z_n]$; $K(z,Z) = [K(z,z_1), K(z,z_2), \ldots, K(z,z_n)]$; $K(Z,z) = K(z,Z)^\top$; and $K(Z,Z) = [K(z_i,z_j)]_{i,j}$. With :
\begin{equation}
    K(z,z') = \exp\left(-\frac{\lVert z - z'\rVert_2^2}{2\tau^2}\right)
\end{equation}

\paragraph{RBF PB} We now consider a VAE with a stochastic generator. That means that it now outputs a distribution $\mathcal{X} \xrightarrow[g_\sigma]{g_\mu} \gZ \xrightarrow[f_\sigma]{f_\mu}$. We thus have that $z = g_\mu(x) + \varepsilon g_\sigma(x)$ with $\varepsilon \sim \mathcal{N}(0, I)$ and $\hat{x} = f_\mu(z) + \varepsilon f_\sigma(z)$ with $\varepsilon \sim \mathcal{N}(0, I)$.

Such networks are trained in two steps. First we train $g_\mu$,$g_\sigma$ and $f_\mu$ classically using the ELBO loss. Then we freeze $g_\mu$,$g_\sigma$ and $f_\mu$ and train $f_\sigma$ using the same ELBO loss. The difference with a regular VAE is that instead of using $p(x|z) = \mathcal{N}(f_\mu(z), I)$, we now use $p(x|z) = \mathcal{N}(f_\mu(z), f_\sigma(z))$ as a prior.

The issue is that even if we take the pullback of such networks, we still don't enforce manifold support. Consequently,  specific design for $f_\sigma$ is needed to enforce that the variance is small on the manifold and large off the manifold. \citet{arvanitidis2018latent} proposes to use a RBF network for $f_\sigma$. That is we instead learn $\beta_\psi = \frac{1}{f_\sigma^2}$ where:
\begin{equation}
    \beta_\psi(z) = W v(z) + \zeta \quad \text{with} \quad v_k(z) = \exp\left(-\frac{\lVert z-z_k\rVert^2_2}{2\tau_k^2}\right).
\end{equation}
Where we use $K$ centroids and positive weights $W$ and bias $\zeta$. Note that we can plug this function, coined \textit{RBF Layer}, as a the first layer of the decoder and then add other layers to learn a more complex function. In any cases, we need to ensure that all of the weights are positive to ensure that $\beta_\psi(z) > 0$ for all $z$.

The metric is then defined as follows:
\begin{equation}
    G(z) = J_{f_\mu}(z)^\top J_{f_\mu}(z) + J_{f_\sigma}(z)^\top J_{f_\sigma}(z).
\end{equation}

\paragraph{PBIG} In \citet{arvanitidis2022pulling}, the authors adopt an information geometry point of view. As the decoder now outputs a distribution, its output is a statistical manifold naturally equipped with the Fisher-Rao metric. In this paper we only consider that the decoder outputs a multivariate diagonal Gaussian distribution. We consequently have direct access to the Fisher-Rao metric of the model and don't need to rely on the KL divergence "black box tools" used in their paper. We still need to have a correctly calibrated estimate of the variance, here we follow the same design as for the \textbf{RBF PB} metric. The metric is thus now given by
\begin{equation}
    G(z) = J_{f_\mu}(z)^\top\text{diag}\left(\frac{1}{f_\sigma(z_i)^2}\right) J_{f_\mu}(z) + J_{f_\sigma}(z)^\top\text{diag}\left(\frac{2}{f_\sigma(z_i)^2}\right) J_{f_\sigma}(z).
\end{equation}

\subsection{Remark on the pullback metrics}
Computing the pullback metric is computationally very expensive and scales poorly with the size of the decoder network. To this end we avoid materializing the full dense jacobian as much as possible. When needed, we use finite differences approximation as they prove faster and less expensive than true autodiff because they only require two forward passes. That is, using $e_j$ the j-th standard basis vector and $\varepsilon$ as small constant, we have that the j-th column of the jacobian matrix is given by:
\begin{equation}
    J_f(z)[ :, j] = \frac{f(z+\varepsilon e_j) - f(z-\varepsilon e_j)}{2\varepsilon}.
\end{equation}
Otherwise the computation of the norm and dot products can be computed using regular Jacobian vector dot products as implemented in automatic differentiation libraries as : $\langle u, v\rangle_{G_{pb}(z)} = \langle u, J^\top(z)J(z)v\rangle = \langle J(z)u, J(z)v\rangle$. In practice, apart from toy networks, this is still too expensive and we resort to materializing the full jacobian matrix using finite differences to compute dot products.

\subsection{Parameter selection}
All of the presented above metrics require to tune some hyper-parameters.

We denoted by $\lambda$ the hyper-parameter regularizing towards the identity. It is needed to avoid numerical errors and ensure a that the metric yields PSD matrices when evaluated. In practice it is chosen in all experiments as $\lambda=10^{-3}\ll1$.

Then these metrics are evaluated on a subset of $K$ centroids that we extract using K-Medoids \citep{schubert2022fast}. To tune the temperature parameters, we build on the approach described in \citet{arvanitidis2018latent} and others. We first introduce a new hyper-parameter $a\in\mathbb{R}^+_*$. For $k\in\llbracket 1,K\rrbracket$, we consider the temperature $\tau_k$ associated to the centroids $z_k$. Denote by $C_k$ the cluster of points whose closest centroids is $z_k$. We thus have the following quantities :
\begin{align}
    r_k &= \text{clamp}\left( \frac{\mid C_k\mid -1}{\text{min\_c\_size}-1}, 0, 1\right)\\
    \text{scale}_k^2 &= r_k \times\text{local\_scale}_k + (1-r_k) \times \text{global\_scale}_k \\
    \tau_k^2 &= a\times \text{quantile\_clip}(\text{scale}_k^2, 0.25,0.95)
\end{align}

The reliability score $r_k$ allows to interpolate between a local scale and a more global scale. This enables to avoid the degenerated cases of very small clusters or high values of $K$. We also clip the values of $\text{scale}_k$ based on the values of the quantiles to further avoid outliers. The value of $\text{local\_scale}_k$ is given similarly to \citet{arvanitidis2018latent} :
\begin{equation}
    \text{local\_scale}_k = \frac{1}{\mid C_k \mid} \sum_{z_i \in C_k} \lVert z_i -z_k \rVert^2 \,.
\end{equation}

And the $\text{global\_scale}_k$ is given by the distance squared to the $n_\text{rank}$ closest other centroid.

In practice, we pick $\text{min\_c\_size}=3$, $n_\text{rank} = 5$, and $a=1$ for all metrics except for \textbf{CovCentroids} and \textbf{PBCentroids} where we set $a=5$.

The temperature of the Gaussian Process for the Warped Pullback metric \citet{sun2025geometry} is given by the maximum distance to the second closest centroid.

We find that these heuristics provide consistent results without much of the hassle of tuning these parameters. In particular it doesn't yield large gap in the values of the local probability density $\sqrt{\det G(z)}$ between training and validation samples. We nevertheless acknowledge that it would be of interest to study these heuristics properties in further details.

\section{Wasserstein gradient flow learning}
\label{app:wasserstein}
We briefly present the method of \citet{huizing2026stories} with their parameter $\alpha$ set to $0$. We consider that the data are sampled according to a distribution $\mu_s$. We model the evolution of $\mu_s$ as a Wasserstein gradient flow for a potential energy $E$. That is for a sample $z$, its Euclidean gradient flow is given by 

\begin{equation}
    \label{eq:wassertstein_flow}
    \frac{\mathrm{d}z}{\mathrm{d}s} = -\nabla E(z).
\end{equation}

As we don't have access to the true potential $E$, we approximate it using a neural network $E_\theta$. More precisely, using the discretized distribution $(\mu_{s_k})_k$, we aim to match $\mu_{s_{k+1}}$ and $(Id - \tau\nabla E_\theta)_{\#} \mu_{s_k}$. We compare the distribution using the debiased Wasserstein distance $\mathcal{W}_\varepsilon$ \citep{feydy2019interpolating}. Hence the loss is :
\begin{equation}
    \mathcal{L}(\theta) = \sum_k (s_{k+1} - s_k) \mathcal{W}_\varepsilon \left( \mu_{s_{k+1}}, (Id - \tau\nabla E_\theta)_{\#} \mu_{s_k} \right)
\end{equation}
At training time we use $\tau = s_{k+1} - s_k$. At inference, we sample an initial point $(z,s)$ and an arrival $s_\text{max}>s$. We then compute the extrapolated trajectory by integrating Eq.\ref{eq:wassertstein_flow} with $E_\theta$ where $\tau = \frac{s_\text{max} - s}{T-1}$ ; where $T$ is the desired number of time points. In the Shepp-Logan dataset setup, we discretize the distribution by binning the sequential factors in 10 bins of equal size.

\section{Geodesics computations}
\label{app:geodesic_comp}

Geodesics are computed using a two stages approach. For the Riemannian case, first a weighted graph is computed from the embedded data. The edges are computed from a KNN graph using 8 neighbors. The edge $e_{i\rightarrow j}$ between $z_i$ and $z_j$ is weighted by the length of $\gamma_{i\rightarrow j}(t) = t z_i + (1-t)z_j$ under the metric. At inference we connect the point to the graph using again the length of the straight line to the closest points under the metric $G$. For the Finsler case, the graph is now directed and we can have different weights between $e_{i\rightarrow j}$ and $e_{j\rightarrow i}$, where the distance is computed using $g$. This yields a first geodesic approximation using Dijkstra's algorithm. For sufficiently dense uniform random graph, this first approximation converges towards the true geodesic \citep{kimDeepGenerativeGeodesics2024a}.

We then proceeded differently for diagonal or dense metrics. Dense metrics induce prohibitive computational cost that prevent using sophisticated algorithms. This is further discouraged on pullback metric where the optimization process might need to differentiate through the pull-backed tensor, inducing additional instabilities.

For the diagonal metrics, we refine the first geodesic approximation from the graph approach using the GEORCE algorithm from \citet{rygaardGEORCEFastNew2025}.

For dense metrics, we use the first geodesic approximation from the graph approach to define a B-spline. We then optimize its control points with the Adam optimizer ($n_{\text{epochs}} = 500, lr=0.01$) by minimizing the curve's energy.

These different approaches enable to have comparable runtime when evaluating different frameworks.

\section{Ablations}
\label{app:ablations}
In this section we investigate several choice of parameter or parametrization done in the main body of the paper; namely, the polar decomposition of the \textit{1-form} $\omega$, the robustness of the framework to noisy labels, and the hyper-parameters of the extrapolation. 

\subsection{Randers metric parametrization}
Here we verify that the use of the polar decomposition of $\omega$ from Eq.\ref{eq:omega_polar_form} is useful. To check this we compare the setting where $\rho_\theta$ is learned or fixed to $1$. We display the results in \tabref{tab:ablation_polar_form}. The polar decomposition doesn't appear to significantly increase performances. This is actually to be expected as the accuracy of the learned hyperplane is already really close to $1$ across the latent space for all base metric. For we have on the test set that $\rho_\theta(z) = 0.873 \pm 0.009$ for the \textit{LAND} metric or even $0.9999 \pm 0.0004$ for the \textit{LAND RBF} metric. We nonetheless expect that this decomposition could prove useful in some settings.

{%
\setcitestyle{numbers,square,citesep={,}} 
\begin{table}[ht]
    \centering
    \caption{Ablation study on the amplitude parametrization. Here, $d=64, \alpha=5.0, \eta=0.999$, and $\zeta=0.9$.}
    \label{tab:ablation_polar_form}
    \begin{tabular}{lccccc}
    \toprule
        & & \multicolumn{2}{c}{Interpolation} & \multicolumn{2}{c}{Extrapolation} \\
        \cmidrule(lr){3-4} \cmidrule(lr){5-6}
        Base metric & Learned $\rho_\theta$ & Kendall Tau ($\uparrow$) & FWD ($\downarrow$) & Kendall Tau ($\uparrow$) & Relative Variation (\%) \\
        \midrule
    \multirow{2}{*}{Euclidean} & \xmark & $1.0 \pm 0.02$ & $0.41$ & $1.0 \pm 0.02$ & $85.59 \pm 33.52$ \\
     & \checkmark & $1.0 \pm 0.02$ & $0.41$ & $1.0 \pm 0.02$ & $85.59 \pm 33.52$ \\
\midrule
    \multirow{2}{*}{LAND \citep{arvanitidis2016locally}} & \xmark & $0.46 \pm 0.29$ & $0.19$ & $1.0 \pm 0.0$ & $90.00 \pm 44.26$ \\
     & \checkmark & $0.46 \pm 0.29$ & $0.19$ & $1.0 \pm 0.0$ & $90.00 \pm 44.26$ \\
\midrule
    \multirow{2}{*}{LAND RBF \citep{kapusniak2024metric}} & \xmark & $0.46 \pm 0.3$ & $0.22$ & $1.0 \pm 0.02$ & $83.47 \pm 30.76$ \\
     & \checkmark & $0.46 \pm 0.3$ & $0.22$ & $1.0 \pm 0.02$ & $83.47 \pm 30.76$ \\
\midrule
    \multirow{2}{*}{CovCentroids \citep{chadebec2022geometric}} & \xmark & $0.45 \pm 0.29$ & $0.2$ & $1.0 \pm 0.03$ & $74.53 \pm 74.71$ \\
     & \checkmark & $0.45 \pm 0.29$ & $0.21$ & $1.0 \pm 0.04$ & $83.33 \pm 80.28$ \\
\midrule
    \multirow{2}{*}{PBCentroids} & \xmark & $0.77 \pm 0.31$ & $0.23$ & $0.59 \pm 0.55$ & $28.75 \pm 70.48$ \\
     & \checkmark & $0.63 \pm 0.39$ & $0.24$ & $0.37 \pm 0.66$ & $15.32 \pm 65.99$ \\
\bottomrule
\end{tabular}
\end{table}
}%

\subsection{Noise robustness}
We investigate how robust our framework is to noise. Consequently we consider a corrupted dataset $\mathcal{D}=\{x_i,z_i,\tilde{s}_i\}$ where the sequential factors are corrupted. More specifically we have that $\tilde{s}_i = s_i + c\times \varepsilon $ with $c$ the label noise parameter, $\varepsilon\sim\mathcal{N}(0,\sigma^2)$, and $\sigma^2$ the empirical variance of the training sequential factors.

The results in \tabref{tab:ablation_label_noise} demonstrate that the method is robust to small noise perturbation. For the interpolation task, the monotonicity score maintains an approximately constant level for $c<1$. It degrades for higher values. In the extrapolation setting, albeit the Kendall Tau remains high, the latent space actually traversed drops quickly as demonstrated by the sharp decrease in relative variation. 

{%
\setcitestyle{numbers,square,citesep={,}}
\begin{table}[ht]
    \centering
    \caption{Ablation study on the anoise robustness. Here, $d=64, \alpha=5.0, \eta=0.999$, and $\zeta=0.9$.}
    \label{tab:ablation_label_noise}
    \begin{tabular}{lccccc}
    \toprule
        & & \multicolumn{2}{c}{Interpolation} & \multicolumn{2}{c}{Extrapolation} \\
        \cmidrule(lr){3-4} \cmidrule(lr){5-6}
        Base metric & Label Noise & Kendall Tau ($\uparrow$) & FWD ($\downarrow$) & Kendall Tau ($\uparrow$) & Relative Variation (\%) \\
        \midrule
    \multirow{2}{*}{Euclidean} & 0.0 & $0.95 \pm 0.24$ & $0.4$ & $1.0 \pm 0.02$ & $85.81 \pm 32.90$ \\
     & 0.1 & $0.98 \pm 0.14$ & $0.35$ & $1.0 \pm 0.01$ & $91.93 \pm 44.27$ \\
     & 0.5 & $0.89 \pm 0.38$ & $0.38$ & $0.99 \pm 0.08$ & $55.37 \pm 25.70$ \\
     & 1.0 & $0.67 \pm 0.69$ & $0.41$ & $1.0 \pm 0.03$ & $30.65 \pm 20.89$ \\
     & 2.0 & $0.44 \pm 0.86$ & $0.41$ & $0.94 \pm 0.27$ & $13.43 \pm 7.88$ \\
     & 5.0 & $0.08 \pm 0.96$ & $0.35$ & $0.65 \pm 0.64$ & $5.84 \pm 5.72$ \\
\midrule
    \multirow{2}{*}{LAND \citep{arvanitidis2016locally}} & 0.0 & $0.44 \pm 0.3$ & $0.21$ & $1.0 \pm 0.03$ & $88.98 \pm 42.75$ \\
     & 0.1 & $0.47 \pm 0.3$ & $0.23$ & $1.0 \pm 0.03$ & $94.08 \pm 49.42$ \\
     & 0.5 & $0.42 \pm 0.34$ & $0.18$ & $1.0 \pm 0.04$ & $76.00 \pm 28.60$ \\
     & 1.0 & $0.31 \pm 0.41$ & $0.21$ & $1.0 \pm 0.05$ & $82.99 \pm 53.65$ \\
     & 2.0 & $0.21 \pm 0.45$ & $0.2$ & $0.99 \pm 0.11$ & $47.10 \pm 26.35$ \\
     & 5.0 & $0.05 \pm 0.49$ & $0.19$ & $0.98 \pm 0.16$ & $23.01 \pm 9.94$ \\
\midrule
    \multirow{2}{*}{LAND RBF \citep{kapusniak2024metric}} & 0.0 & $0.49 \pm 0.3$ & $0.25$ & $1.0 \pm 0.02$ & $85.89 \pm 33.61$ \\
     & 0.1 & $0.42 \pm 0.31$ & $0.19$ & $1.0 \pm 0.02$ & $85.49 \pm 34.21$ \\
     & 0.5 & $0.41 \pm 0.34$ & $0.18$ & $1.0 \pm 0.01$ & $59.06 \pm 21.57$ \\
     & 1.0 & $0.31 \pm 0.4$ & $0.22$ & $0.99 \pm 0.11$ & $34.98 \pm 12.43$ \\
     & 2.0 & $0.21 \pm 0.45$ & $0.22$ & $0.92 \pm 0.36$ & $16.54 \pm 7.85$ \\
     & 5.0 & $0.05 \pm 0.5$ & $0.2$ & $0.75 \pm 0.59$ & $9.39 \pm 8.15$ \\
\midrule
    \multirow{2}{*}{CovCentroids \citep{chadebec2022geometric}} & 0.0 & $0.43 \pm 0.32$ & $0.17$ & $0.48 \pm 0.56$ & $21.30 \pm 82.48$ \\
     & 0.1 & $0.45 \pm 0.32$ & $0.22$ & $0.33 \pm 0.59$ & $9.75 \pm 87.14$ \\
     & 0.5 & $0.41 \pm 0.33$ & $0.16$ & $0.83 \pm 0.39$ & $81.04 \pm 115.09$ \\
     & 1.0 & $0.32 \pm 0.41$ & $0.15$ & $0.79 \pm 0.43$ & $74.41 \pm 109.27$ \\
     & 2.0 & $0.21 \pm 0.45$ & $0.18$ & $0.67 \pm 0.53$ & $48.50 \pm 77.56$ \\
     & 5.0 & $0.04 \pm 0.49$ & $0.19$ & $0.6 \pm 0.55$ & $36.92 \pm 84.42$ \\
\midrule
    \multirow{2}{*}{PBCentroids} & 0.0 & $0.65 \pm 0.33$ & $0.16$ & $0.95 \pm 0.25$ & $64.57 \pm 32.54$ \\
     & 0.1 & $0.63 \pm 0.34$ & $0.17$ & $1.0 \pm 0.05$ & $65.43 \pm 25.77$ \\
     & 0.5 & $0.68 \pm 0.37$ & $0.19$ & $0.98 \pm 0.16$ & $57.61 \pm 27.85$ \\
     & 1.0 & $0.52 \pm 0.52$ & $0.17$ & $0.93 \pm 0.31$ & $32.47 \pm 30.76$ \\
     & 2.0 & $0.34 \pm 0.63$ & $0.16$ & $0.94 \pm 0.22$ & $13.75 \pm 13.48$ \\
     & 5.0 & $0.11 \pm 0.69$ & $0.14$ & $0.91 \pm 0.28$ & $12.50 \pm 15.93$ \\
\bottomrule
\end{tabular}
\end{table}
}%

\subsection{Extrapolation hyper-parameters}
\begin{figure*}[ht]
\begin{center}
\includegraphics[width=\linewidth]{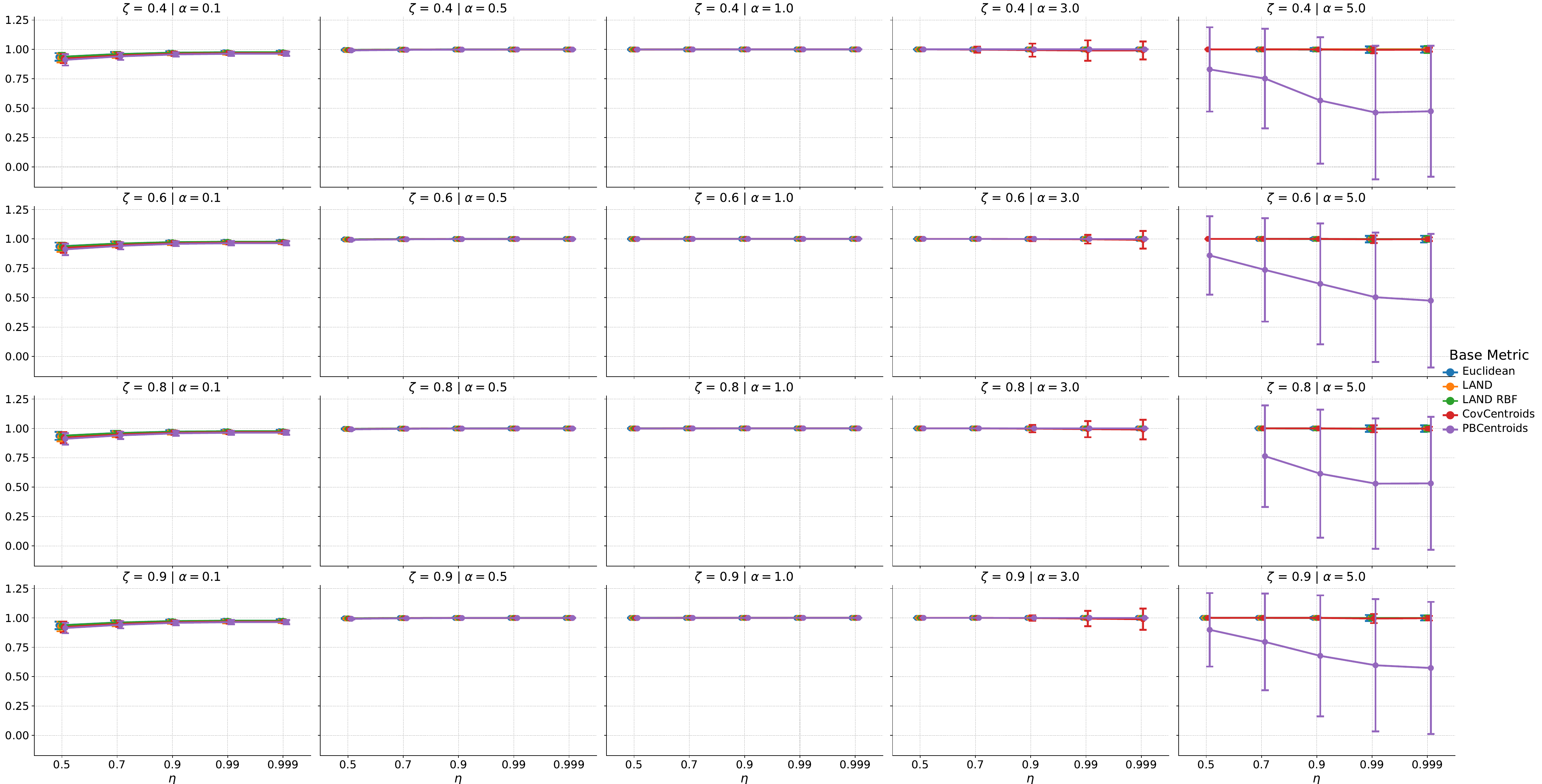}
\end{center}
\caption{\textbf{Monotonicity of extrapolated trajectories}: Evaluation of the extrapolation framework across different hyper-parameters. Here the latent dimension is $64$. The error bar correspond to the standard deviation across the computed trajectories with 25 time points.}
\label{fig:shepplogan_extrapolation_kendalltau_grid}
\end{figure*}

\begin{figure*}[ht]
\begin{center}
\includegraphics[width=\linewidth]{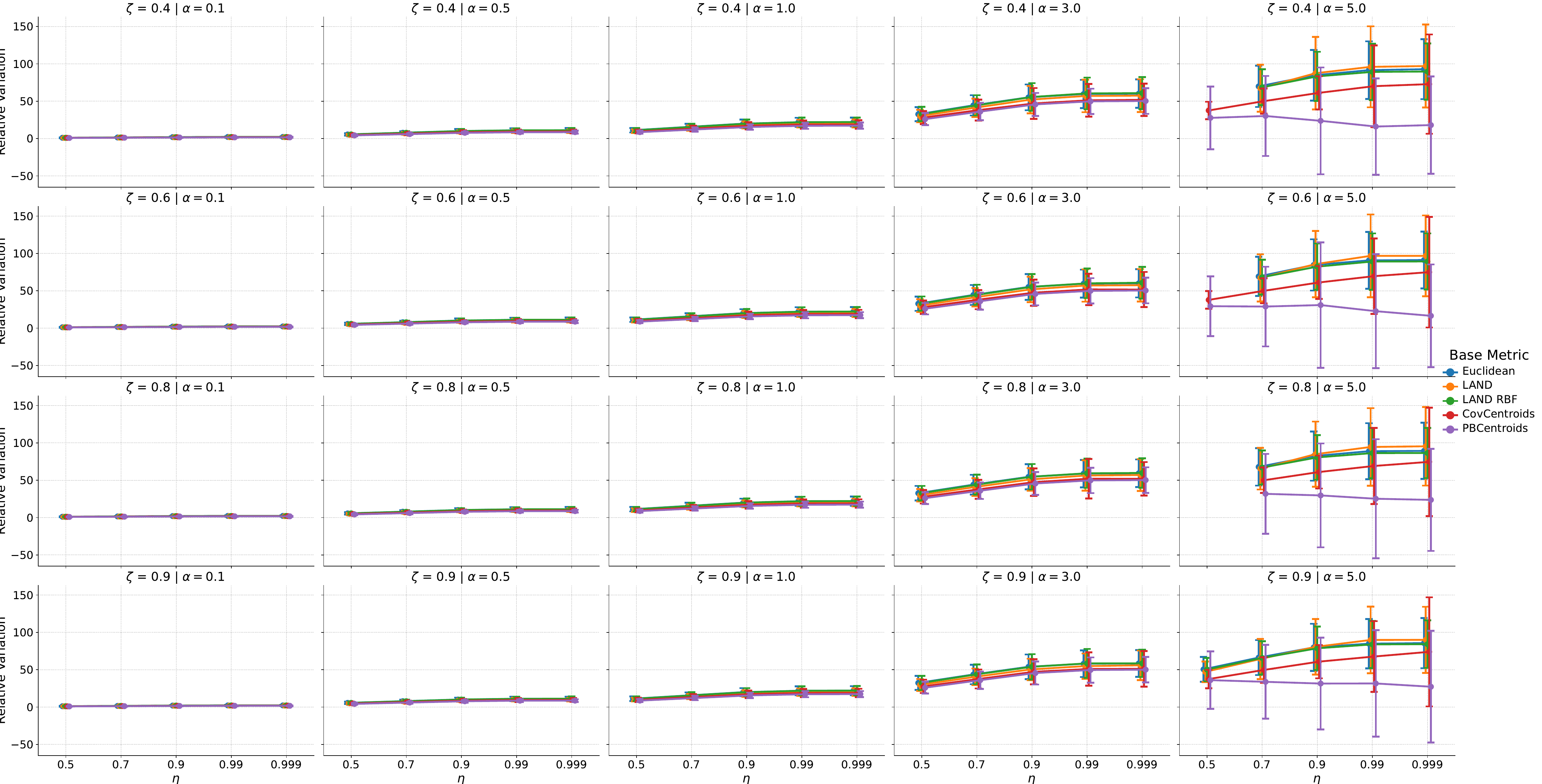}
\end{center}
\caption{\textbf{Relative variation of extrapolated trajectories}: Evaluation of the extrapolation framework across different hyper-parameters. Here the latent dimension is $64$. The error bar correspond to the standard deviation across 500 computed trajectories with 25 time points.}
\label{fig:shepplogan_extrapolation_sizevariation_grid}
\end{figure*}

The extrapolation under Finsler Geometry relies on a correct tuning of the target concentration parameter $\eta$, the norm of the speed vector $\alpha$ and the exponential moving average coefficient $\zeta$. We display in Figure  \ref{fig:shepplogan_extrapolation_kendalltau_grid} and \ref{fig:shepplogan_extrapolation_sizevariation_grid} the variation of the monotonicity score and the relative variation when varying the parameters. We only include the results on asymmetric frameworks. The results demonstrate that the momentum sampling parameter $\zeta$ has little influence on the results. It also demonstrates that even low values of $\eta$ yield sequentially consistent trajectories, but to obtain a better traversal of the latent space, a higher value of $\alpha$ is needed.

\section{Concentration parameter tuning}
\label{app:kappa_vmf_extrapolation}

In the context of extrapolation, we need to carefully tune the choice of the initial speed. As we designed the metric so that the expected direction of evolution is in the direction of $-\omega$, we sample accordingly using a von Mises Fisher distribution. To calibrate its concentration parameter, we aim to choose it so that we control the average cosine similarity between the sampled speed and the direction of the vector field $\omega$.
Let $x\sim \text{vMF}(\mu,\kappa)$ for some unit vector $\mu\in\mathbb{S}^{d-1}$ the mean direction and concentration parameter $\kappa\in\mathbb{R}^+$. For some target value $\eta$, we aim to find $\kappa$ such that $\eta = \mathbb{E}_x[\text{cosim}(\mu,x)]$.
We denote $f$ the density of the distribution such that :
\begin{align}
    f(x) &= C_d(\kappa) \exp(\kappa \mu^\top x) \\
    \text{with }\, C_d(k) &= \frac{\kappa^{\frac{d}{2}-1}}{\left(2\pi\right)^{\frac{d}{2}}I_{\frac{d}{2}-1}(\kappa)}
\end{align}
Here $I$ is the modified Bessel function of the first kind at order $\frac{d}{2}-1$.
As both, $\lVert x\rVert_2=1$ and $\lVert \mu\rVert_2=1$, the cosine similarity between $x$ and $\mu$ is given by $\text{cosim}(\mu,x)=\mu^\top x$.

We make use of the log partition trick. Note:
\begin{equation}
    Z(\kappa) = \frac{1}{C_d(\kappa)} = \int_{\mathbb{S}^{d-1}} \exp(\kappa \mu^\top x) \mathrm{d}\sigma(x)
\end{equation}
Then because :
\begin{equation}
    \frac{\mathrm{d}Z}{\mathrm{d}\kappa} = \int_{S^{d-1}} \mu^\top x \exp(\kappa \mu^\top x) \mathrm{d}\sigma(x)
\end{equation}

And $\log(C_d) = \frac{d-2}{2}\log(\kappa) - \log I_{\frac{d-2}{2}}(\kappa) - \frac{d}{2}\log(2\pi)$, we have the expected cosine similarity:
\begin{align}
    \mathbb{E}[\mu^\top x] &= Z(\kappa) \frac{\mathrm{d}Z}{\mathrm{d}\kappa} \\
    &=\frac{\mathrm{d}\log Z}{\mathrm{d}\kappa} = -\frac{\mathrm{d}\log C_d}{\mathrm{d}\kappa} \\
    &= \frac{2 -d}{2\kappa} + \frac{I'_{\frac{d}{2}-1}(\kappa)}{I_{\frac{d}{2}-1}(\kappa)} \\
    &= \frac{2 -d}{2\kappa}  + \frac{
    I_{\frac{d}{2}}(\kappa) + \frac{d-2}{2\kappa} I_{\frac{d}{2}-1}
    }{
    I_{\frac{d}{2}-1}(\kappa)
    } \\
    &= \frac{I_{\frac{d}{2}}(\kappa)}{I_{\frac{d}{2}-1}(\kappa)}
\end{align}
Where we used the fact that $I'_p(v) = \frac{p}{v}I_p(v) + I_{p+1}(v)$.

Consequently for a target cosine similarity $\eta\in[0,1]$, we solve, for $\kappa$, $\eta =\frac{I_{\frac{d}{2}}(\kappa)}{I_{\frac{d}{2}-1}(\kappa)}$. Following the work of \citet{banerjee2005clustering}, we use the approximation of the ratio of Bessel functions so that 
\begin{equation}
    \label{eq:kappa_by_eta}
    \kappa = \frac{\eta d - \eta^3}{1-\eta^2}
\end{equation}

\section{Experimental details}
\label{app:experimental_details}
Unless mentioned otherwise we use the same set of hyper-parameters for each experiment. For more details, consult the available source code. The Shepp-Logan images are $256\times256$ black and white images in the range $[0,1]$.

\textbf{Encoder} The encoder downsamples the input through five convolution blocks: Conv($1\to16$), Conv($16\to32$), Conv($32\to64$), Conv($64\to128$), and Conv($128\to256$). The first four blocks use stride-$2$ convolutions with $3\times3$ kernels, reducing spatial dimensions from $256\times256$ to $16\times16$. The final block uses stride-$1$ to preserve spatial resolution. Each block applies instance normalization with affine parameters and PReLU activation. The resulting $256\times16\times16$ feature map is flattened and projected through separate linear layers to produce the mean $\mu$ and log-variance $\log\sigma^2$ of the latent distribution.

\textbf{Decoder} The decoder maps the latent code back to image space through a symmetric upsampling path. A linear layer projects the latent vector to $256\times16\times16$ dimensions, followed by five upsampling blocks that restore the spatial resolution to $256\times256$. Each upsampling block consists of nearest-neighbor $2\times$ upsampling, a $3\times3$ convolution, instance normalization (without affine parameters), and PReLU activation. The final layer applies a $3\times3$ convolution followed by sigmoid activation to produce the reconstruction.

\textbf{Randers metric} Both the direction and amplitude of the Randers metric $\omega$ are MLP with two layers and SiLU activations. The amplitude regressor part is fed through a sigmoid layer to ensure that the results stay in $[0,1]$, hence keeping the positivity of the metric.

\textbf{Regressor} In the Shepp-Logan experiments, for the regressor we use a two stage approach. First we directly segment the image using morphological operations to retrieve the top left ellipsis. Once segmented, we can fit an ellipsis and then compute its area. However when the geodesics cross regions of the latent space where the decoder yields less well reconstructed images, the morphological segmentation fails. In such cases we train a \textit{resnet18} network to directly predict the value of interest. The morphological segmentation yields more precise results on correctly decoded images, hence this pipeline. 

\textbf{Training recipe} The VAE is trained using a combination of the regular ELBO-loss and the MS-SSIM loss from \citet{zhao2015loss} for improved reconstruction. We train for 3000 epochs with a cosine learning rate schedule. The training and evaluation are done using a single Nvidia Ada 6000.

\section{Pseudo code}
\label{app:algorithms}
In this section we summarize the algorithms needed to learn the Randers metric and to evaluate the interpolated and extrapolated trajectories. Note that the user is free to choose the initial encoder/decoder as well as the base metric $G$. We advise to refer to the code and the paper for the GEORCE geodesic solver \citet{rygaardGEORCEFastNew2025}. For completeness we reproduce the pseudo code for the Riemannian solver.

\begin{algorithm}
    \caption{Learning $\tilde{\phi}_\theta$}
    \label{alg:learn_phi}
    \begin{algorithmic}
        \Require $\{(z_i, s_i)\}_{i=1}^N$ : latent points and their associated sequential factors. $G$ : the Riemannian metric
        \State Initialize $\tilde{\phi}_\theta$ randomly
        \While{not converged}
            \For{Batch of points $\{z_i,s_i\}_{i=1}^B$}
                \For{each $z_i$ in the batch}
                    \State $y_i^j = \sign(s_i - s_j)$ for $z_j \in \text{KNN}(z_i)$ \Comment{Local pseudo-labels}
                    \State $\tilde{l}_i^j =\langle\tilde{\phi}_\theta(z_i),(z_j - z_i)\rangle_{G(z_i)}$ 
                    \State $l_\phi(z_i) = \frac{1}{K_\phi} \sum_{j} \max(0, 1 - y_i^j \times \tilde{l}_i^j)$ \Comment{Local hinge loss}
                \EndFor
                \State Update $\tilde{\phi}_\theta$ using $\frac{1}{B}\sum_{i=1}^Bl_\phi(z_i)$
            \EndFor
        \EndWhile
    \end{algorithmic}
\end{algorithm} 

\begin{algorithm}
    \caption{Learning $\rho_\theta$}
    \label{alg:learn_rho}
    \begin{algorithmic}
        \Require $\{(z_i, s_i)\}_{i=1}^N$ : latent points and their associated sequential factors, $\tilde{\phi}_\theta$ : learned direction field, $G$ : the Riemannian metric
        \State Initialize $\rho_\theta$ randomly
        \While{not converged}
            \For{Batch of points $\{z_i,s_i\}_{i=1}^B$}
                \For{each $z_i$ in the batch}
                    \State $y_i^j = \text{sign}(s_j - s_i)$ for $z_j \in \text{KNN}(z_i)$ \Comment{Local pseudo-labels}
                    \State $\tilde{l}_i^j =\langle\tilde{\phi}_\theta(z_i),(z_j - z_i)\rangle_{G(z_i)}$ 
                    \State $\text{acc}_\phi(z_i) = \frac{1}{K} \sum_{j} \mathrm{1}[y_i^j\times \tilde{l}_i^j >0]$ \Comment{Hyperplane accuracy}
                    \State $l_\rho(z_i) = \lVert \rho_\theta(z_i) - 2\times \max\left(\text{acc}_\phi(z_i) - 0.5\right) \rVert^2$ \Comment{Regression loss}
                \EndFor
                \State Update $\rho_\theta$ using $\frac{1}{B}\sum_{i=1}^Bl_\rho(z_i)$
            \EndFor
        \EndWhile
    \end{algorithmic}
\end{algorithm}

\begin{algorithm}
    \caption{Randers metric construction}
    \label{alg:randers_construction}
    \begin{algorithmic}
        \Require $(z,v)\in T\gZ$ : point and tangent vector, $\tilde{\phi}_\theta$ : learned direction field, $\rho_\theta$ : learned amplitude regressor, $G$ : base metric, $\epsilon$ : small positive constant to ensure positivity
        \State $\phi_\theta(z) = \frac{\tilde{\phi}_\theta(z)}{\lVert \tilde{\phi}_\theta(z)\rVert_{G^{-1}(z)} + \epsilon}$ \Comment{Normalize direction field}
        \State $\omega_\theta(z) = \rho_\theta(z) \times \phi_\theta(z)$  \Comment{Polar form of $\omega$}
        \State $\FF(z,v) = \sqrt{v^TG(z)v} + \langle \omega_\theta(z), v\rangle$ \Comment{Randers metric definition}
    \end{algorithmic}
\end{algorithm}

\begin{algorithm}
    \caption{Geodesic evaluation}
    \label{alg:evaluation_geodesic}
    \begin{algorithmic}
        \Require $\{(z_i, s_i)\}_{i=1}^N$ : latent points and their associated sequential factors, $\FF$ : learned Finsler-Randers metric, $D$ : decoder, $r$ : regressor to retrieve sequential factors from images
        \For{each pairs $(z_0, z_1)$ such that $s_0 < s_1$}
            \State $\gamma_t = \text{Geodesic}(\FF, z_0, z_1)$ \Comment{Compute geodesic between $z_0$ and $z_1$ under metric $\FF$}
            \State $\hat{x}(t) = D(\gamma_t)$ \Comment{Decode the geodesic to get images}
            \State $\hat{s}(t) = r(\hat{x}(t))$ \Comment{Retrieve sequential factors from the decoded images}
            \State Compute $m(\hat{s}(t))$ \Comment{Compute monotonicity score of the retrieved sequential factors}
        \EndFor
    \end{algorithmic}
\end{algorithm}

\begin{algorithm}
    \caption{GeneralizedImplicitLeapfrog}
    \label{alg:leapfrog}
    \begin{algorithmic}
        \Require $H$ : hamiltonian function to integrate, $(q_0,p_0)$ : initial position and momentum, $L$: number of integration step, $\gamma$ : step size.
        \State $k=0$
        \State $q_k,p_k = q_0,p_0$
        \While{$k < L$}
            \State $p_{k+\frac{1}{2}} = p_k - \gamma \nabla_q H(q_k,p_{k+\frac{1}{2}}) $
            \Comment{Half update momentum (implicit equation)}
            \State $q_{k+1} = q_k + \frac{\gamma}{2}\left(\nabla_p H(q_k, p_{k+\frac{1}{2}}) + \nabla_p H(q_{k+1},p_{k+\frac{1}{2}}) \right)$
            \Comment{Update position (implicit equation)}
            \State $p_{k+1} = p_{k+\frac{1}{2}} - \gamma \nabla_q H (q_{k+1},p_{k+\frac{1}{2}}) $
            \Comment{Update momentum}
            \State $k = k+1$
        \EndWhile
    \end{algorithmic}
\end{algorithm}

\begin{algorithm}
    \caption{Extrapolation computation}
    \label{alg:computation_extrapolation}
    \begin{algorithmic}
        \Require $z_0\in\gZ$ : start latent point, $\FF^*$ : learned dual Finsler-Randers metric, $\omega$: learned vector field, $\eta$ : target cosine similarity, $\alpha$ : norm of the initial speed, $\zeta$ : EMA parameter of the momentum, $T$ : number of points.
        
        \State $\kappa = \frac{\eta d - \eta^3}{1-\eta^2}$ 
        \Comment{Eq. \ref{eq:kappa_by_eta}} 
        \State $q_t = z_0$
        \State $p_t = 0$
        \For{ $t\in \llbracket 1, T \rrbracket$}
            \State $v_t = \alpha\times \text{vMF}(-\frac{\omega(q_t)}{\lVert\omega(q_t)\rVert_2},\kappa)$ \Comment{Sample initial speed}
            \State $\tilde{p}_t = g(q_t, v_t) v_t$
            \Comment{Legendre transform}
            \State $p_t = (1-\zeta) \tilde{p}_t + \zeta p_t$
            \Comment{Exponential Moving Average of the momentum}
            \State $q_t,p_t = $ GeneralizedImplicitLeapfrog$(\frac{1}{2}\FF_*^2,q_t,p_t,L=1,\gamma=\frac{1}{T})$
        \EndFor
    \end{algorithmic}
\end{algorithm}

\begin{algorithm}[!ht]
    \caption{GEORCE for Riemannian manifolds}
    \label{alg:georce}
    \begin{algorithmic}[1]
        \State \textbf{Input}: $\mathrm{tol}$, $T$
        \State \textbf{Output}: Geodesic estimate $x_{0:T}$
        \State Set $x_{t}^{(0)} \leftarrow a+\frac{b-a}{T}t$, $u_{t}^{(0)} \leftarrow \frac{b-a}{T}$ for $t=0.,\dots,T-1$ and $i \leftarrow 0$
        \While{$\lVert \nabla_{y}E(y)\mid_{y=x_{t}^{(i)}}\rVert_{2} > \mathrm{tol}$}
        \State $G_{t} \leftarrow G\left(x_{t}^{(i)}\right)$ for $t=0,\dots,T-1$
        \State $\nu_{t} \leftarrow \nabla_{y}\left(u_{t}^{(i)}G\left(y\right)u_{t}^{(i)}\right)\mid_{y=x_{t}^{(i)}}$ for $t=,1\dots,T-1$
        \State $\mu_{T-1} \leftarrow \left(\sum_{t=0}^{T-1}G_{t}^{-1}\right)^{-1}\left(2(a-b)-\sum_{t=0}^{T-1}G_{t}^{-1}\sum_{t>j}^{T-1}\nu_{j}\right)$
        \State $\tilde{u}_{t} \leftarrow -\frac{1}{2}G_{t}^{-1}\left(\mu_{T-1}+\sum_{j>t}^{T-1}\nu_{j}\right)$ for $t=0,\dots,T-1$
        \State Using line search find $\alpha^{*}$ for the following optimization problem with the discrete energy functional $E$
        \begin{equation*}
            \begin{split}
                \alpha^{*} = \argmin_{\alpha}\quad &E\left(\tilde{x}_{0:T}\right) \quad \text{(exact line search)} \\
                \text{s.t.} \quad &\tilde{x}_{t+1}=\tilde{x}_{t}+\alpha \tilde{u}_{t}+(1-\alpha)u_{t}^{(i)}, \quad t=0,\dots,T-1, \\
                &\tilde{x}_{0}=a.
            \end{split}
        \end{equation*}
        \State Set $u_{t}^{(i+1)} \leftarrow \alpha^{*}\tilde{u}_{t}+(1-\alpha^{*})u_{t}^{(i)}$ for $t=0,\dots,T-1$
        \State Set $x_{t+1}^{(i+1)} \leftarrow x_{t}^{(i+1)}+u_{t}^{(i+1)}$ for $t=0,\dots,T-1$
        \State $i \leftarrow i+1$
        \EndWhile
        \State return $x_{t}$ for $t=0,\dots,T-1$
    \end{algorithmic}
\end{algorithm}


\newpage
\section{Interpolation and Extrapolation example}

\begin{figure*}[ht]
\begin{center}
\includegraphics[width=\linewidth]{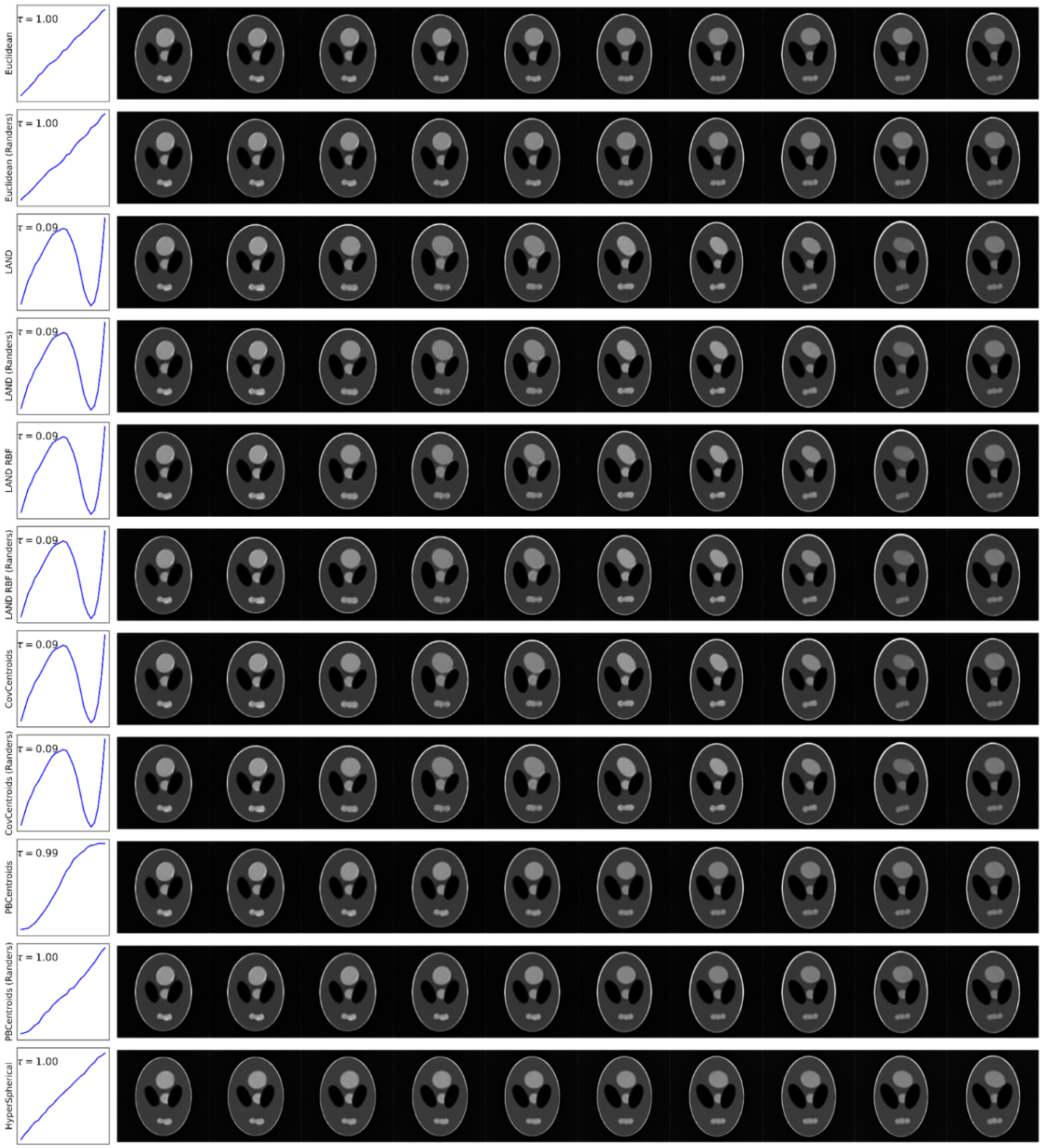}
\end{center}
\caption{Example of interpolation with different framework with fixed endpoints on the Shepp-Logan dataset . Left columns show the predicted sequential factors along the trajectory.}
\label{fig:interpolation_examples_shepplogan}
\end{figure*}

\begin{figure*}[ht]
\begin{center}
\includegraphics[width=\linewidth]{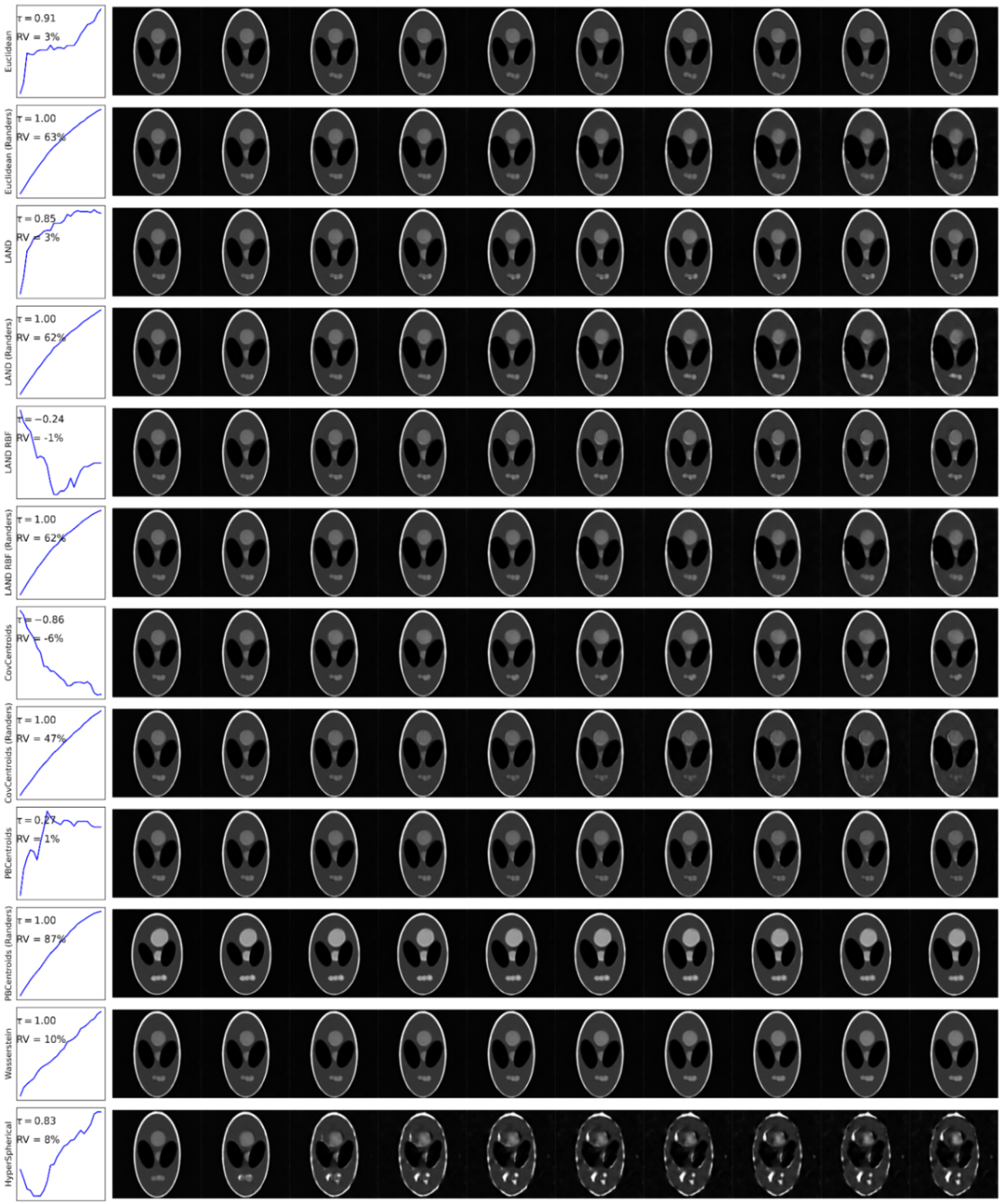}
\end{center}
\caption{Example of extrapolation with different framework on the Shepp-Logan dataset. Left columns show the predicted sequential factors along the trajectory. Here we have $\alpha=5.0$, $\zeta=0.9$, and $\eta=0.999$}
\label{fig:extrapolation_examples_shepplogan}
\end{figure*}

\begin{figure*}[ht]
\begin{center}
\includegraphics[width=\linewidth]{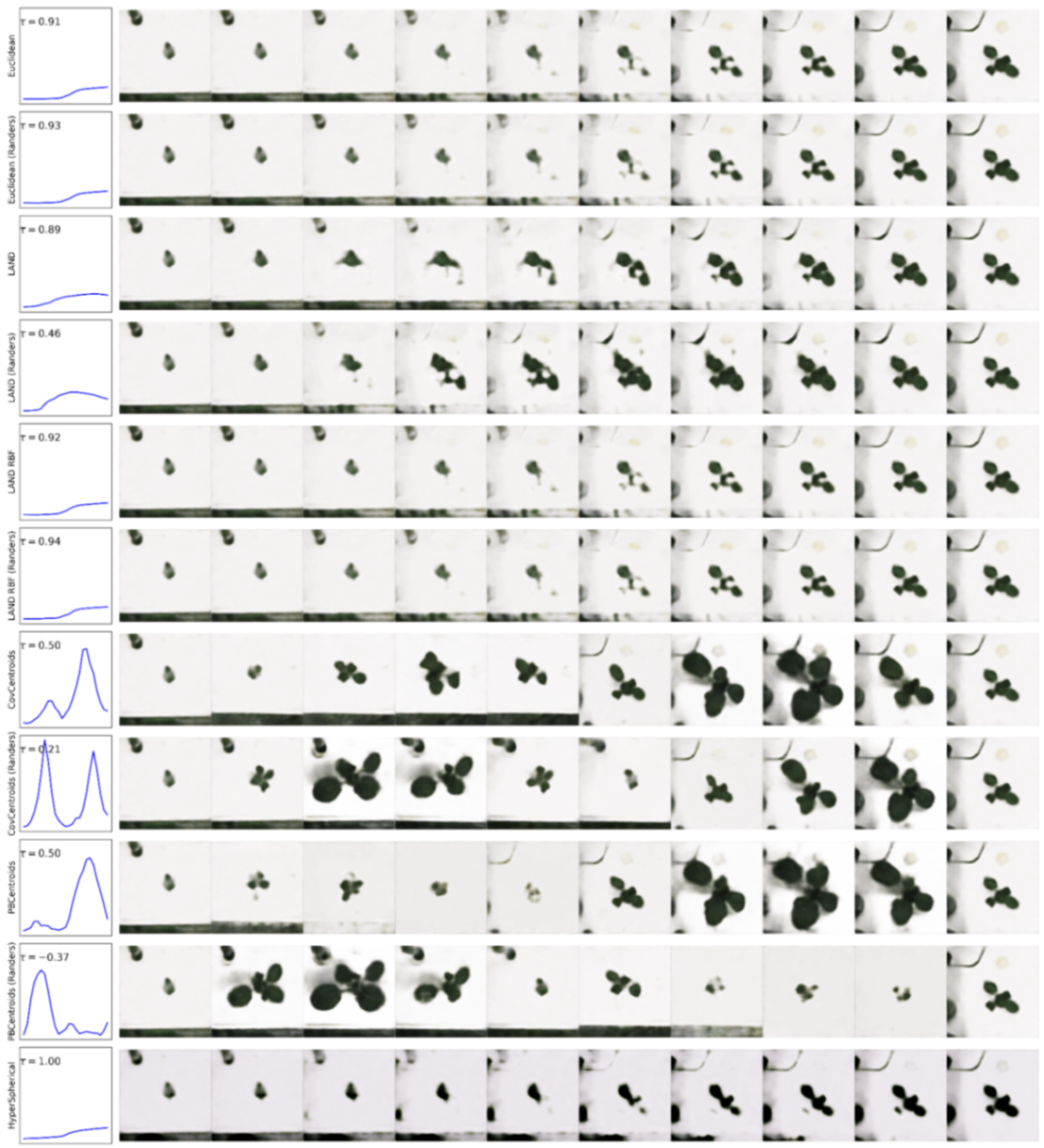}
\end{center}
\caption{Example of interpolation with different framework with fixed endpoints on the KOMATSUNA dataset . Left columns show the predicted sequential factors along the trajectory.}
\label{fig:interpolation_examples_komatsuna}
\end{figure*}

\begin{figure*}[ht]
\begin{center}
\includegraphics[width=\linewidth]{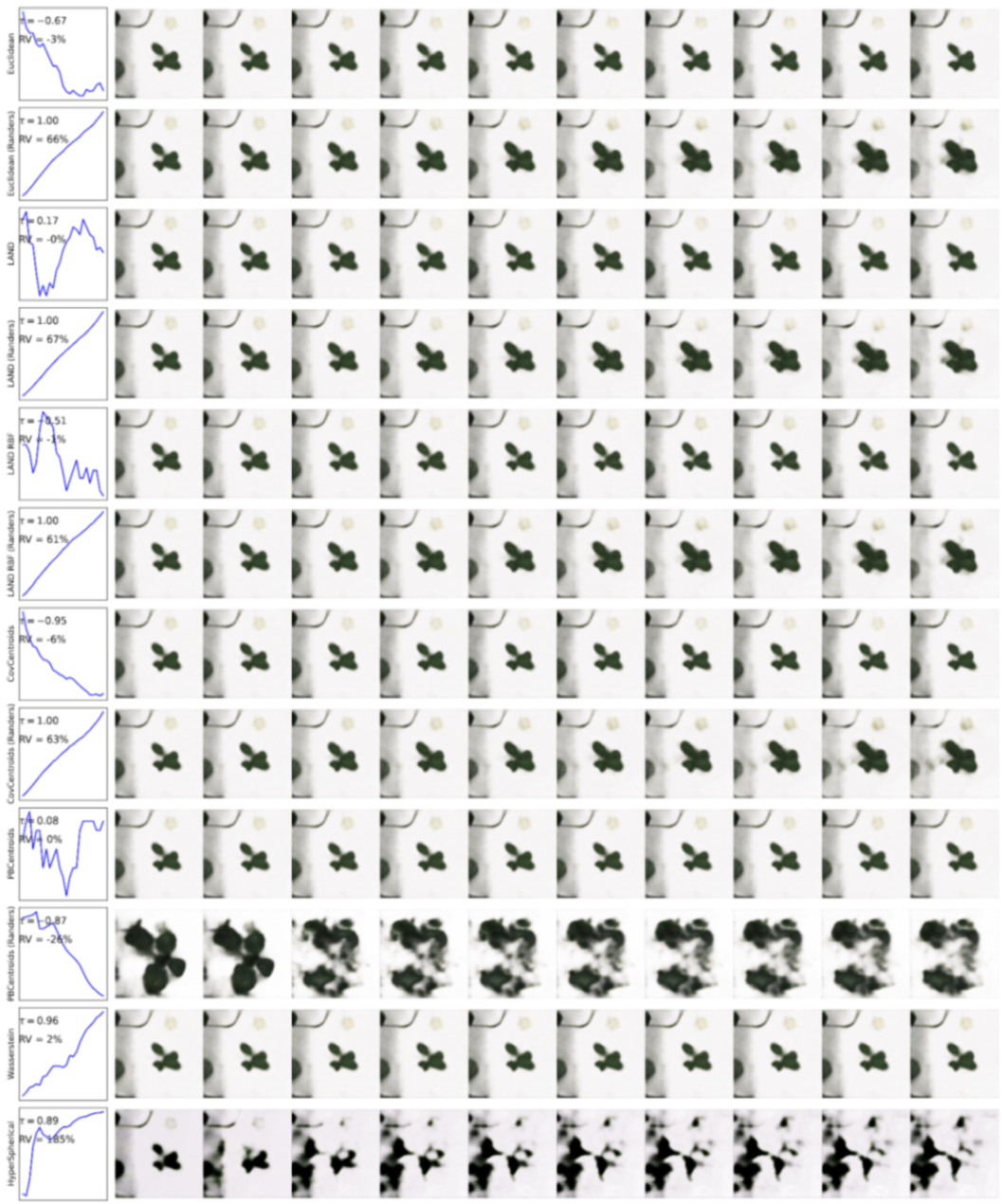}
\end{center}
\caption{Example of extrapolation with different framework on the KOMATSUNA dataset. Left columns show that predicted sequential factors along the trajectory. Here we have $\alpha=5.0$, $\zeta=0.9$, and $\eta=0.999$}
\label{fig:extrapolation_examples_komatsuna}
\end{figure*}

\end{document}

%% file: math_commands.tex
\usepackage{amsmath,amsfonts,bm}
\usepackage{pifont}    %

\newcommand{\xmark}{\ding{55}}%
\newcommand{\FF}{\mathcal{F}}

\def\eqref#1{equation~\ref{#1}}

\def\1{\bm{1}}

\DeclareMathAlphabet{\mathsfit}{\encodingdefault}{\sfdefault}{m}{sl}
\SetMathAlphabet{\mathsfit}{bold}{\encodingdefault}{\sfdefault}{bx}{n}

\def\gD{{\mathcal{D}}}

\def\gP{{\mathcal{P}}}

\def\gX{{\mathcal{X}}}

\def\gZ{{\mathcal{Z}}}

\newcommand{\R}{\mathbb{R}}

\DeclareMathOperator*{\argmin}{arg\,min}

\DeclareMathOperator{\sign}{sign}

\usepackage{amsthm}
\usepackage{thmtools}
\usepackage[capitalise,noabbrev]{cleveref}

\newtheorem{definition}{Definition}
\newtheorem{proposition}{Proposition}

\crefname{definition}{Definition}{Definitions}
\Crefname{definition}{Definition}{Definitions}
\crefname{proposition}{Proposition}{Propositions}
\Crefname{proposition}{Proposition}{Propositions}

\newcommand{\appref}[1]{Appendix~\ref{#1}}
\newcommand{\tabref}[1]{Table~\ref{#1}}